\documentclass[journal]{IEEEtran}
\usepackage[hidelinks]{hyperref}
\usepackage{amssymb}
\usepackage{mathtools}
\usepackage{times}
\usepackage{graphicx}
\usepackage{subfigure}
\usepackage{multirow}
\usepackage{multicol}
\usepackage{amsmath,lipsum}
\usepackage{cuted}
\usepackage{bm}
\allowdisplaybreaks

\usepackage{dashrule}
\usepackage{algorithm}
\usepackage{algpseudocode}
\usepackage[table]{xcolor}
\usepackage{threeparttable}
\usepackage{rotating}
\usepackage{dcolumn}
\usepackage{booktabs}
\usepackage{hyperref}
\usepackage{bigstrut}

\usepackage{amsthm}

\usepackage{latexsym}
\usepackage{textcomp}
\usepackage{cite}
\usepackage{colortbl}
\definecolor{light-gray}{gray}{0.75}

\usepackage[normalem]{ulem}
\newcommand\hl{\bgroup\markoverwith
	{\textcolor{yellow}{\rule[-.5ex]{2pt}{2.5ex}}}\ULon}

\usepackage{mathptmx}
\DeclareMathAlphabet{\mathcal}{OMS}{cmsy}{m}{n}
\DeclareMathOperator*{\argmin}{arg\,min}

\usepackage{epstopdf}
\usepackage{lineno}
\usepackage{xcolor}

\begin{document}
%\linenumbers
\title{Algorithm Modelization: Transforming Optimization Algorithms into Evolvable Intelligence}

\author{
    Jian~Yao,
    Zhiyao~Zhang,
    Yichen~Li,
    Xingyu~Wu\textsuperscript{$\dagger$},
    and~Kay~Chen~Tan\textsuperscript{$\dagger$},~\IEEEmembership{Fellow,~IEEE}%
    \thanks{
        Jian Yao, Zhiyao Zhang, Yichen Li, Xingyu Wu,
        and Kay Chen Tan are with the Department of
        Data Science and Artificial Intelligence,
        The Hong Kong Polytechnic University,
        Hong Kong SAR, China.
        Jian Yao and Zhiyao Zhang contributed equally to this work.
        \textsuperscript{$\dagger$}Corresponding authors:
        Kay Chen Tan (kctan@polyu.edu.hk)
        and Xingyu Wu (xingy.wu@polyu.edu.hk).
    }
}
\IEEEtitleabstractindextext{%
\begin{abstract}
Can algorithms be learned and evolved as models, rather than repeatedly redesigned as programs? We explore this question by introducing Algorithm Modelization, a new paradigm that turns iterative algorithms into trainable models. Its central idea is to regard algorithm execution trajectories as computational experience: by learning how algorithmic states evolve over iterations, a model can internalize the knowledge embodied in existing algorithms, reproduce their computational behaviors, and further evolve beyond them through training. As a first realization of this paradigm, we modelize iterative optimization as an autoregressive process, where preceding generations constitute the computational context for generating the next population. This perspective allows trajectories produced by fundamentally different optimizers to be expressed and learned within a unified model. We develop a dimension-independent population-temporal architecture to capture both population interactions and iterative dynamics, together with an imitation-to-surpassing learning framework that first acquires algorithmic knowledge from heterogeneous teachers and then improves the resulting optimizer directly through optimization feedback. The modelized optimizer consequently goes beyond reproducing individual algorithms: it consolidates heterogeneous search knowledge into a single trainable model, achieves stronger optimization capability through subsequent improvement, and generalizes its acquired knowledge across problems and substantially larger unseen dimensions. More fundamentally, Algorithm Modelization makes algorithmic capability accumulable through learning, where new algorithms and problem-solving experience can become training resources for progressively improving the model. Our results establish a promising path from manually designed algorithms toward general algorithm models that learn, generalize, and evolve from computational experience.
\end{abstract}

\begin{IEEEkeywords}
Algorithm Modelization; Trajectory Imitation;  Online Learning; Evolvable Intelligence; Optimization
\end{IEEEkeywords}
}

\maketitle
\IEEEdisplaynontitleabstractindextext
\IEEEpeerreviewmaketitle
\section{Introduction}
\label{sec:introduction}
Algorithms are fundamental building blocks of computational intelligence, yet their improvement has traditionally relied on repeated human redesign~\cite{10494009, liu2024evolution}. Over decades, increasingly powerful algorithms have been developed by introducing new heuristics, update rules, and computational mechanisms~\cite{brest2006self,11638110}, but the knowledge accumulated during this process remains largely embedded in manually specified procedures~\cite{9321762,11200037}. When a stronger algorithm emerges, it typically replaces or modifies its predecessors rather than directly learning from and refining their computational experience~\cite{11493956}. Modern machine learning~\cite{goodfellow2016deep} suggests a fundamentally different possibility: knowledge can be encoded in trainable models, accumulated from data, and continuously improved through further training. This raises a natural question: can iterative algorithms themselves be transformed into trainable models that inherit from existing algorithms and subsequently improve beyond them?

We formulate this problem as Algorithm Modelization. Given trajectories generated by iterative algorithms, Algorithm Modelization aims to internalize the computational behavior manifested through their state transitions into a trainable neural model, which captures how computational states evolve over iterations and remains trainable for further performance improvement. Unlike conventional predictive models that learn task-level input-output mappings, Algorithm Modelization learns the iterative update behavior embodied in algorithm trajectories, replacing fixed update procedures with a trainable model that directly drives the optimization process. Since the procedural knowledge of an iterative algorithm is manifested through how its computational states evolve during execution, Algorithm Modelization seeks to encode these transition behaviors into model parameters, enabling them to be learned, generalized, and continuously refined through training.

In this work, we investigate Algorithm Modelization through iterative optimization as a representative application. Optimization algorithms naturally generate sequential trajectories in which search states progressively evolve toward better solutions~\cite{11475039}. Inspired by autoregressive sequence modeling~\cite{vaswani2017attention}, we formulate this iterative process as learning to generate the next optimization state conditioned on its recent history. Given the latest $K$ generations, the model extracts the evolutionary relationships contained in the trajectory and generates the next population. Conceptually, previous optimization states form the computational context, analogous to preceding tokens in language modeling, while the next population corresponds to the next state to be generated. By expressing optimizer executions through a common population-transition representation, this formulation provides a unified interface for trajectories originating from substantially different optimization paradigms. In this study, evolutionary, Bayesian, and gradient-based optimizers~\cite{9804806,ament2023unexpected,chen2023symbolic} are represented through this common interface, allowing heterogeneous search behaviors to be learned within the same modelization framework.

Algorithm Modelization of optimization trajectories presents several fundamental challenges. First, a population trajectory simultaneously contains population structure and temporal structure: individuals interact within each generation, while their search behaviors evolve across successive generations. Capturing both forms of dependency is essential for understanding how optimization progresses~\cite{hong2026enhancing}. Second, algorithmic models should preserve their computational form across different problem dimensions, so that learned search knowledge can transfer beyond the dimensional configurations observed during training~\cite{li2024pretrained}. Third, although different optimizers can be expressed through a common population-transition interface, their trajectories still contain heterogeneous search behaviors that must be encoded into a shared learnable representation capable of preserving transferable transition knowledge. Finally, Algorithm Modelization involves not only learning demonstrated behaviors but also developing the model's own optimization capability, requiring a training mechanism that further refines inherited knowledge according to optimization performance.

To address these challenges, we develop a history-conditioned model architecture for optimization algorithm modelization. Historical populations are aligned to preserve individual-level correspondence across generations, and each individual is represented by relative fitness information together with a dimension-independent Coordinate-Set representation that describes population geometry through shared coordinate processing and symmetric aggregation~\cite{zaheer2017deep}. A population-temporal architecture then jointly models interactions among individuals and temporal evolution across generations~\cite{vaswani2017attention,lee2019set}, followed by a future-state decoder that infers the search state for the next iteration. The predicted representation is converted into population evolution through a signed dense interaction operator and coordinate-level crossover, followed by objective-based selection to obtain the next population. The overall architecture therefore maps recent optimization history directly to subsequent population evolution and provides a unified neural representation of iterative search dynamics.

%We further develop a two-stage training strategy corresponding to the two central capabilities required by Algorithm Modelization: imitation and surpassing. During the first stage, trajectories generated by teacher algorithms provide structured supervision for learning population evolution. The training data combine search experience from heterogeneous optimization paradigms, allowing the model to acquire diverse algorithmic behaviors through a unified trajectory representation. During the second stage, the model performs optimization with its learned policy and is further trained according to the quality of the resulting search process. Differentiable meta-training propagates optimization-performance signals through the learnable population-generation mechanism, enabling the model to refine and reorganize the computational knowledge acquired during trajectory imitation.
%As a result, algorithm trajectories provide rich computational experience from which effective search behaviors can be acquired, while subsequent performance-driven training further enhances the resulting optimizer. Our main contributions are summarized as follows:

We further develop a two-stage training strategy built around the two capabilities that distinguish Algorithm Modelization from either algorithm distillation or learning an optimizer from scratch: imitation and surpassing. Imitation is necessary because the purpose of modelization is to inherit the computational knowledge already embodied in effective algorithms, rather than discarding this accumulated experience and relearning search behavior solely from optimization feedback. Yet imitation alone is insufficient for our goal: training only to reproduce demonstrated transitions provides no explicit objective for improving optimization performance beyond the demonstrated behavior, reducing modelization primarily to behavioral imitation rather than an independently improvable algorithmic model. Conversely, directly optimizing a randomly initialized model provides no mechanism for inheriting the structured search knowledge accumulated by existing algorithms. We therefore view imitation and surpassing as two complementary stages of a single learning process: the former provides an algorithmic prior from existing computational experience, while the latter turns inherited knowledge into an independently improvable search capability.

During the first stage, trajectories generated by teacher algorithms provide structured supervision for learning population evolution. The training data combine search experience from heterogeneous optimization paradigms, allowing the model to acquire diverse algorithmic behaviors through a unified trajectory representation. During the second stage, teacher supervision is removed and the model interacts with optimization problems under the state distribution induced by its own search behavior. On differentiable training functions, performance feedback is propagated through the learnable population-generation mechanism, allowing the model to refine the computational knowledge acquired during trajectory imitation. After training, the resulting optimizer is deployed with fixed parameters and requires only function evaluations. This imitation-to-surpassing process changes the role of existing algorithms from fixed procedures to reusable sources of computational experience: their knowledge provides the starting point of learning, rather than the upper bound of the resulting model.

Our main contributions are summarized as follows:

\begin{enumerate}
\item Conceptually, we introduce Algorithm Modelization, a new paradigm that turns iterative algorithms from fixed computational procedures into trainable models. We formulate algorithm execution trajectories as a new form of algorithmic data and introduce a general learning paradigm in which models acquire iterative computational behavior from existing algorithms and can be further improved through training. We instantiate this paradigm in optimization, opening a direction for algorithm development in which computational knowledge can be learned, accumulated, and improved through models.

\item Technically, we establish an autoregressive modelization framework that enables optimization algorithms to be learned from their iterative trajectories and subsequently surpassed. We formulate optimization as next-population generation conditioned on the preceding generations and develop a dimension-independent population-temporal architecture to capture both search interactions and evolutionary dynamics. On top of this architecture, we introduce an imitation-to-surpassing learning scheme that first internalizes computational knowledge from heterogeneous teacher algorithms and then directly improves the acquired search capability through performance-driven meta-training. These designs provide an end-to-end technical realization of Algorithm Modelization, from trajectory representation and behavior acquisition to autonomous performance improvement.

\item Methodologically, Algorithm Modelization yields a new class of optimization methods that are powerful, generalizable, reusable, and extensible through learning. The resulting optimizer can consolidate complementary search knowledge from fundamentally different algorithmic paradigms and further improve, providing a path toward stronger optimization performance without repeatedly redesigning search rules by hand. Its dimension-independent formulation enables direct extrapolation to substantially larger unseen dimensions, while the same modelization architecture and learning framework can be instantiated across different optimization settings without scenario-specific architectural redesign. The acquired search capability can also be reused under new deployment conditions by coupling the learned optimizer with task-specific adaptive modules, rather than relearning the underlying search policy from scratch. More importantly, its capability is extensible: newly available algorithms can continuously provide additional computational experience for training and progressively strengthen the existing model, while broader problem classes and optimization scenarios may further serve as future sources of computational experience. This provides a promising path toward increasingly general optimization models whose capabilities can accumulate and evolve through learning.

\end{enumerate}

%The remainder of this paper is organized as follows. Section~\ref{sec:related} reviews related work. Section~\ref{sec:method} presents the proposed Algorithm2Model framework. Section~\ref{sec:experiments} reports the experimental evaluation, and Section~\ref{sec:con} concludes the paper.

\section{Related Work}
\label{sec:related}

Iterative optimization algorithms have attracted broad attention across scientific and engineering applications. Representative paradigms include population-based evolutionary optimization, which evolves a set of candidates through variation and selection~\cite{das2010differential}; Bayesian optimization, which selects points for fitness evaluations (FEs) by combining a probabilistic surrogate with an acquisition function~\cite{11311712}; and gradient-based optimization, which updates solutions using local derivative information, as exemplified by Lion~\cite{chen2023symbolic}. Despite their different search mechanisms, these optimizers all generate iterative computational trajectories, while the knowledge governing their search behavior remains encoded in fixed procedures rather than accumulated from execution experience~\cite{9321762}.

To move beyond fixed procedures and obtain optimizers that adapt and generalize across problems, existing research has followed three main routes. The first is large language model (LLM) based automated algorithm design, which represents an optimizer as code and searches for improved programs through generation, evaluation, and iterative refinement~\cite{11686185}. EoH~\cite{liu2024evolution} evolves LLM-generated heuristics according to task performance, MEoH~\cite{yao2025multi} extends this process to multi-objective algorithm design, and ReEvo~\cite{ye2024reevo} introduces reflective evolution for refining candidate heuristics. These methods can produce effective algorithms for a target scenario, but their central operation is to generate a new program. They do not directly absorb the execution trajectories of established optimizers into a persistent model, so the procedural knowledge embodied in existing algorithms is not explicitly inherited and accumulated; when the scenario changes, algorithm design must generally be performed again.

The second route is {Meta-Black-Box Optimization (MetaBBO)}, which formulates optimizer adaptation as a bi-level learning problem over a distribution of tasks~\cite{10993463}. A meta-policy learns to select, configure, or coordinate predefined low-level optimization components. For example, DE-DDQN~\cite{sharma2019deep} selects differential-evolution mutation strategies, whereas LDE~\cite{9359652} adapts control parameters through policy-gradient learning. MetaBBO therefore learns how an optimizer should adapt its handcrafted components to the current search state, but it does not modelize the complete computational behavior of an existing optimizer: the underlying operator space remains specified in advance, and learning is primarily responsible for its control.

The third route is {learning to optimize (L2O)}, which directly parameterizes the optimization mechanism and learns its update behavior from data or optimization feedback~\cite{tang2024learn}. One representative pattern learns search behavior directly from performance signals without inheriting the realized behavior of an existing optimizer. LGA and LES~\cite{lange2023discovering,lange2023discoveringES} replace conventional evolutionary update rules with attention-based population policies, while B2Opt~\cite{li2025b2opt} learns black-box search components under a limited FE budget. POM~\cite{li2024pretrained} learns mutation and crossover strategies end to end by differentiating performance signals through population evolution, and EPOM~\cite{han2026enhancing} follows the same principle while strengthening task-feature modeling, population interaction, and gradient stability. Although effective, these methods initialize their search capability without behavioral knowledge inherited from established optimizers. Consequently, useful search principles already embodied in mature optimizers are not explicitly inherited, but are instead learned from task-level optimization feedback, making the resulting capability more dependent on the coverage and quality of the training-function distribution.

A second L2O pattern inherits search behavior through optimizer trajectories. EvoTF~\cite{lange2024evolution} introduces Evolutionary Algorithm Distillation, training a Transformer to predict the search-distribution updates of teacher evolution strategies. This demonstrates that procedural behavior can be transferred from an algorithm into model parameters, but its standard distillation objective is designed primarily to reproduce teacher updates rather than improve upon them. This work also explores meta-evolution and self-referential distillation as possible routes beyond imitation, but reports overfitting in the former and instability in the latter; consequently, it does not establish a stable imitation-to-surpassing pipeline. MetaDistill~\cite{han2026textttmetadistill} subsequently adopts a pretraining--fine-tuning scheme: it first distills trajectories from diverse classical optimizers into a common distributional representation and then optionally performs self-supervised fine-tuning on an unseen target task. Although fine-tuning enables target-specific adaptation, it makes improved deployment problem dependent: each new problem requires fresh observations and an additional model-update process. Repeating this adaptation for every target introduces extra data, computation, and implementation overhead, rather than allowing one fixed optimizer checkpoint to be applied directly across problems.

Algorithm Modelization builds on this progression by combining inheritance from algorithm execution trajectories with subsequent performance-driven refinement. A2M represents the full population history as an autoregressive computational context, allowing trajectories from evolutionary, Bayesian, and gradient-based optimizers to be learned through a unified interface. It then makes performance-driven improvement an integral second training stage, so that inherited behavior becomes the starting point rather than the endpoint of learning. This design distinguishes A2M from LLM-based methods that redesign programs and MetaBBO methods that learn controllers over predefined operators. Within L2O, it addresses both limitations identified above: unlike POM and EPOM, A2M does not learn search capability entirely from scratch; unlike EvoTF, it explicitly advances from trajectory imitation to performance-based enhancement. Moreover, this enhancement is completed before deployment, rather than repeated as target-specific fine-tuning for every new problem. The resulting fixed A2M checkpoint can therefore inherit established optimization knowledge, improve it through learning, and then be reused directly across unseen problem instances.

\begin{figure*}[h]
\begin{center}
\includegraphics[width=\linewidth]{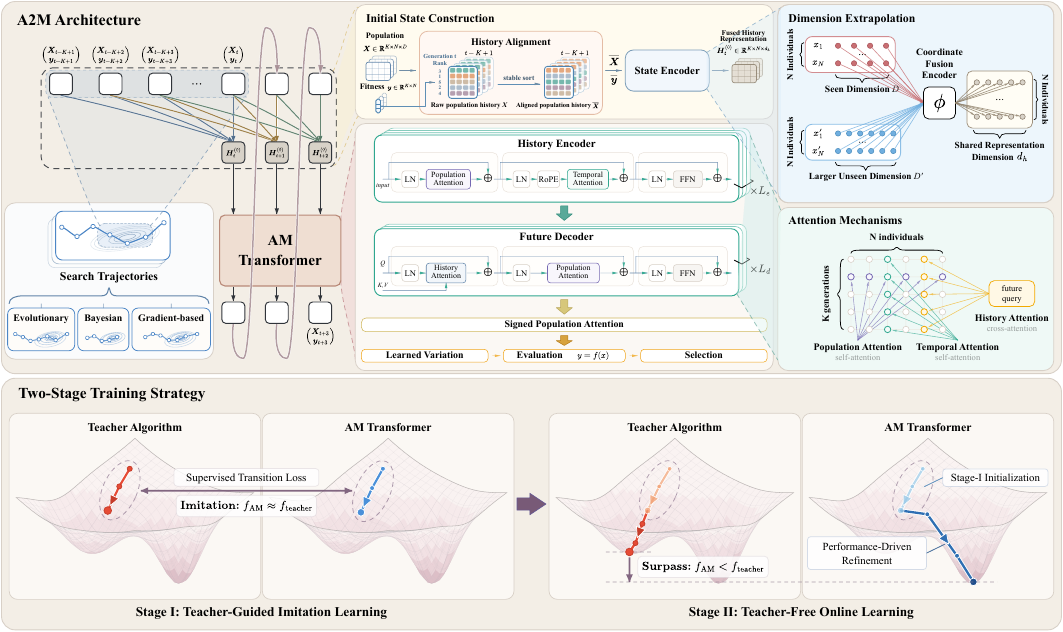}
\end{center}
\caption{Overview of Algorithm Modelization for optimization. The upper part presents the A2M architecture for modeling search trajectories, while the lower part shows the two-stage training strategy from teacher-guided imitation to teacher-free refinement beyond the teacher.}
\label{fig:framework}
\vskip -1em
\end{figure*}

\section{Algorithm-to-Model}
\label{sec:method}
This section presents Algorithm2Model (A2M), our instantiation of
algorithm modelization for population-based black-box optimization.
A2M replaces a fixed handcrafted update rule with a trainable population
transition model conditioned on a short search history. It is first
initialized from teacher trajectories and then refined through
differentiable online learning.
We instantiate A2M in three single-objective optimization settings:
continuous optimization, mixed-variable optimization, and expensive
continuous optimization. The continuous and mixed-variable settings
share the same model architecture and learning framework, differing only
in the training trajectories and evaluation interface. For expensive
optimization, the trained continuous A2M is kept fixed as a candidate
generator and is coupled with an independent X-to-Y branch that ranks the
generated candidates under a limited FE budget. Throughout this paper, we use A2M to refer to both the overall framework and the resulting optimization model, with the intended meaning determined by context.

\subsection{Problem Formulation}
\label{sec:problem_formulation}

We consider a box-constrained single-objective black-box optimization
problem
\begin{equation}
    \bm{x}^{\star}\!
    \in
    \argmin_{\bm{x}\in\Omega_D} f(\bm{x}),
\end{equation}
where \(\Omega_D\subset\mathbb{R}^{D}\) denotes the box-constrained search
domain and objective values are available only through FEs.
At generation \(t\), the optimizer maintains an evaluated population
\(\bm{X}_t\in\mathbb{R}^{N\times D}\) and its fitness vector
\(\bm{y}_t\in\mathbb{R}^{N}\). A2M conditions the next transition on the
most recent \(K\) population states:
\begin{equation}
\left\{
\begin{aligned}
    \mathcal{H}_t
    &=
    \!\Bigl\{
    \!
    \bigl(\bm{X}_{t-K+1},\bm{y}_{t-K+1}\bigr),
    \ldots,
    \bigl(\bm{X}_t,\bm{y}_t\bigr)
    \Bigr\},\\[0.5ex]
    \bm{U}_t
    &=
    \pi_{\theta}\!(\mathcal{H}_t),
\end{aligned}
\right.
\label{eq:a2m_transition}
\end{equation}
The model directly generates a complete
candidate population \(\bm{U}_t\); it does not predict objective values
or perform gradient descent on the decision variables.

The formulation above defines the common population-transition interface used by A2M. For mixed-variable optimization, A2M still maintains and updates a continuous genotype. A deterministic phenotype map \(\mathcal{P}_{q,D}\) rounds a predefined subset of coordinates before each FE:
\begin{equation}
    \widetilde{f}_{q,D}(\bm{x})
    =
    f_q\!\bigl(\,\mathcal{P}_{q,D}(\bm{x})\bigr).
\label{eq:mixed_objective}
\end{equation}
For continuous problems, \(\mathcal{P}_{q,D}\) is the identity map.
Therefore, continuous and mixed-variable problems share the same model
interface, while all selection decisions are based on the corresponding
true objective values.

\subsection{A2M Architecture}
\label{sec:a2m_architecture}

A2M contains four common components: identity-preserving history
alignment, a dimension-independent state encoder, a two-axis history
model with a future population decoder, and a signed population-attention
decoder followed by learnable crossover and one-to-one selection.

\subsubsection{History Alignment and State Encoding}

To preserve meaningful temporal correspondence across generations, A2M maintains row-wise alignment among successive population states when constructing and updating the search history. This provides the temporal model with a consistent view of how population members evolve over recent generations. During A2M optimization, such correspondence is maintained naturally by strict one-to-one selection, where the $i$-th candidate competes only with the $i$-th parent.
To obtain a consistent fitness-aware ordering, we further reorder the entire history according to the current population. Let $\pi_{\,t}$ denote the stable permutation that sorts the current population by ascending fitness. The same permutation is applied to every state in the history window. Specifically, for \(k=t-K+1,\ldots,t\),
\begin{equation}
\left\{
\begin{aligned}
    \overline{\bm{X}}_{\!k}
    &=
    \bm{P}_{\pi_{\,t}}\cdot\bm{X}_k,
    \\[0.5ex]
    \overline{\bm{y}}_k
    &=
    \bm{P}_{\pi_{\,t}}\cdot\,\bm{y}_k.
\end{aligned}
\right.
\label{eq:history_alignment}
\end{equation}
Here, $\bm{P}_{\pi_{\,t}}$ is the corresponding permutation matrix.
Because the permutation is shared across the complete history window,
the existing row correspondence is retained, while the first row is
anchored to the best individual in the current generation.

Each aligned individual is represented by standardized fitness, its within-population rank, and a dimension-independent description of its decision vector.
Specifically, a shared coordinate network processes three coordinate-wise relations: the normalized coordinate value, its deviation from the population mean on that coordinate, and its deviation from the aligned trajectory of the current-best individual. Symmetric mean and RMS pooling aggregate the coordinate features into a fixed-width representation. Because all coordinate parameters are shared and the aggregation is symmetric, the resulting representation is independent of the decision dimension.
A token-wise gate then fuses the coordinate representation with the fitness features and a learned embedding of the fitness-sorted population row. This produces the initial history tensor
\(\bm{H}^{(0)}\!\in\mathbb{R}^{K\times N\times d_h}\).

\subsubsection{Two-Axis History Encoder and Future Decoder}

The fused history representation $\bm{H}^{(0)}$ is processed alternately along the population and temporal axes to capture within-generation interactions and cross-generation evolution.
For the $\ell$-th encoder block, population self-attention \cite{vaswani2017attention,lee2019set} is first
applied independently to the $N$ individuals at each historical state,
followed by temporal self-attention over the $K$ aligned states of each
individual:
\begin{equation}
\left\{
\begin{aligned}
    \widetilde{\bm{H}}^{(\ell)}_{k,:,:}
    &=
    \bm{H}^{(\ell-1)}_{k,:,:}
    +
    \bm{\alpha}^{(\ell)}_{\text{pop}}
    \odot
    \operatorname{MHA}^{(\ell)}_{\text{pop}}
    \!\Bigl(
        \operatorname{LN}
        \bigl(\bm{H}^{(\ell-1)}_{k,:,:}\bigr)
    \!\Bigr),
    \\[0.5ex]
    \widehat{\bm{H}}^{(\ell)}_{:,i,:}
    &=
    \widetilde{\bm{H}}^{(\ell)}_{:,i,:}
    +
    \bm{\alpha}^{(\ell)}_{\text{time}}\!
    \odot
    \operatorname{MHA}^{(\ell)}_{\text{time}}
    \!\Bigl(
        \operatorname{RoPE}
        \bigl(
            \,
            \operatorname{LN}
            \bigl(\widetilde{\bm{H}}^{(\ell)}_{:,i,:}\bigr)
        \!\bigr)
    \!\!\Bigr),
    \\[0.5ex]
    \bm{H}^{(\ell)}
    &=
    \widehat{\bm{H}}^{(\ell)}
    +
    \bm{\alpha}^{(\ell)}_{\text{ffn}}
    \odot
    \operatorname{FFN}^{(\ell)}
    \!\Bigl(
        \operatorname{LN}
        \bigl(\widehat{\bm{H}}^{(\ell)}\bigr)
    \!\Bigr).
    \label{eq:axial_ffn}
\end{aligned}
\right.
\end{equation}
Here, $k=1,\ldots,K$ indexes historical states and
$i=1,\ldots,N$ indexes aligned individuals.
The learnable vectors
$\bm{\alpha}_{\text{pop}}$,
$\bm{\alpha}_{\text{time}}$, and
$\bm{\alpha}_{\text{ffn}}$
control the residual contribution of the corresponding sublayers.
Relative temporal positions are encoded with RoPE\cite{su2024roformer}, with the current
generation assigned position zero. This alternating structure jointly models population interactions and temporal dynamics while preserving their distinct relational structure.

After the final encoder block, the representation of the current
population is used to initialize the future decoder:
\begin{equation}
    \bm{Q}^{(0)}
    =
    \operatorname{LN}
    \bigl(
        \bm{H}^{(L_e)}_{K,:,:}
    \bigr)
    \in \mathbb{R}^{N\times d_h},
\end{equation}
where $L_e$ denotes the number of encoder blocks.
Each row of $\bm{Q}^{(0)}$ therefore serves as the initial future query
for the corresponding current individual.

The future decoder alternates two forms of interaction. Each future query
first attends to the historical representations of its corresponding
individual to capture its recent evolution. The resulting queries then
interact through population self-attention:
\begin{align}
    ~\widetilde{\bm{Q}}^{(\ell)}_{i,:}
    &=
    \bm{Q}^{(\ell-1)}_{i,:}
    +
    \bm{\beta}^{(\ell)}_{\text{hist}}
    \odot
    \operatorname{MHA}^{(\ell)}_{\text{hist}}
    \!\Bigl(
        \operatorname{LN}\!\bigl(\bm{Q}^{(\ell-1)}_{i,:}\!\bigr)
        ;\,
        \operatorname{LN}\!\bigl(\bm{H}^{(L_e)}_{:,i,:}\!\bigr)
        \!
    \Bigr),
    \notag
    \\[0.5ex]
    \mathllap{%
        \smash{%
            \left\{
            \vphantom{
                \begin{gathered}
                    \widetilde{\bm{Q}}^{(\ell)}_{i,:}\\[0.5ex]
                    \widehat{\bm{Q}}^{(\ell)}\\[0.5ex]
                    \bm{Q}^{(\ell)}
                \end{gathered}
            }
            \right.
        }%
        \mkern-8mu
    }
    ~\widehat{\bm{Q}}^{(\ell)}
    &=
    \widetilde{\bm{Q}}^{(\ell)}
    +
    \bm{\beta}^{(\ell)}_{\text{pop}}
    \odot
    \operatorname{MHA}^{(\ell)}_{\text{pop}}
    \!\Bigl(
        \operatorname{LN}\!\bigl(\widetilde{\bm{Q}}^{(\ell)}\bigr)
    \!\Bigr),
    \label{eq:future_decoder}
    \\[0.5ex]
    ~\bm{Q}^{(\ell)}
    &=
    \widehat{\bm{Q}}^{(\ell)}
    +
    \bm{\beta}^{(\ell)}_{\text{ffn}}
    \odot
    \operatorname{FFN}^{(\ell)}
    \!\!\Bigl(
        \operatorname{LN}\!\bigl(\widehat{\bm{Q}}^{(\ell)}\bigr)
    \!\Bigr).
    \notag
\end{align}
Here, $\operatorname{MHA}(\cdot;\cdot)$ denotes cross-attention, while
$\operatorname{MHA}(\cdot)$ denotes self-attention. After $L_d$ decoder
blocks, $\operatorname{LN}\bigl(\bm{Q}^{(L_d)}\!\bigr)$ provides one future
representation for each individual.

\subsubsection{Signed Population Attention and Crossover}

The decoded future representations are converted into candidate solutions
through signed attention over the current population. For each operator
head $h$, the final future representation of individual $i$ is projected
as a query, while the current representation of individual $j$ is
projected as a key:
\begin{equation}
\left\{
\begin{aligned}
    \bm{q}_{i,h}
    &=
    \tanh\!\Bigl(
        \bm{W}^{Q}_{h}\cdot
        \operatorname{LN}\!\bigl(\bm{Q}^{(L_d)}_{i,:}\!\bigr)
        +
        \bm{b}^{Q}_{h}
    \Bigr),\\
    \bm{k}_{\!j,h}
    &=
    \tanh\!\Bigl(
        \bm{W}^{K}_{h}\cdot
        \bm{Q}^{(0)}_{j,:}
        +
        \bm{b}^{K}_{h}
    \Bigr).
\end{aligned}
\right.
\label{eq:operator_qk}
\end{equation}
Their interaction determines the signed population-attention weight
\begin{equation}
    \omega_{\,i\!j}
    =
    b_{i\!j}\cdot
    \tanh\!\Bigl(
        \sum_{h=1}^{H_o}\!
        \alpha_{\,h}\cdot
        \frac{
            \bm{q}_{i,h}^{\top}\cdot\bm{k}_{\!j,h}
        }{\sqrt{d_o}}
    \Bigr),
\label{eq:signed_population_attention}
\end{equation}
where $H_o$ is the number of operator heads, $\alpha_{\,h}$ is a learnable
coefficient for the $h$-th head, and $b_{i\!j}$ is a stochastic structural
mask with $b_{ii}=1$ that regularizes off-diagonal interactions.

The proposal for individual $i$ is obtained by using the current
population as the values:
\begin{equation}
    \widetilde{\bm{x}}_{t,i}
    =
    \sum_{j=1}^{N}
    \omega_{\,i\!j}\cdot\overline{\bm{x}}_{t,j}.
\label{eq:population_proposal}
\end{equation}
Unlike conventional softmax attention, $\omega_{\,i\!j}$ can take either positive or negative values and is not normalized to sum to one. This removes the restriction to convex combinations of the current population and allows A2M to represent a broader range of population-transition behaviors.

A lightweight crossover network then determines how each proposal is
combined with its corresponding parent. It predicts an
individual-specific crossover score from fitness, rank, and
parent--proposal similarity. A straight-through Hard-Gumbel mechanism \cite{jang2016categorical}
converts this score into a hard coordinate-wise mask
$\bm{M}_t\in\{0,1\}^{N\times D}$ in the forward pass, yielding
\begin{equation}
    \bm{U}_t
    =
    \bigl(\bm{1}-\bm{M}_t\bigr)\odot\overline{\bm{X}}_{\!t}
    +
    \bm{M}_t\odot\widetilde{\bm{X}}_t,
\label{eq:a2m_crossover}
\end{equation}
where $\widetilde{\bm{X}}_t$ collects the proposals defined in
Eq.~\eqref{eq:population_proposal}. The corresponding soft path is
retained during backpropagation, making the crossover differentiable
during training.

The resulting candidates are projected to the search bounds and evaluated
by the true objective function. Each candidate competes only with its
corresponding parent and replaces it under strict fitness improvement.
This one-to-one selection maintains the row correspondence used by the
history model across successive generations.

\subsection{Two-Stage Model Learning}
\label{sec:two_stage_learning}

A2M is trained in two stages to first inherit population-transition behavior from teacher trajectories and then further refine the learned search capability under its own state distribution. Stage I learns from teacher-generated search trajectories, while Stage II removes teacher supervision and continues training through differentiable interaction with optimization tasks induced by A2M's own search behavior.

\subsubsection{Stage I: Teacher-Trajectory Imitation}

A2M is first trained to imitate population transitions collected from teacher optimizers.
For each history window, the teacher provides the
pre-crossover proposal together with the corresponding candidate
transition. These signals supervise both the population operator and the
resulting candidate generation:
\begin{equation}
    \mathcal{L}_{\text{sup}}
    =
    \mathcal{L}_{\text{op}}
    +
    \lambda_{\,\text{cand}}\cdot
    \mathcal{L}_{\text{cand}},
\label{eq:supervised_loss}
\end{equation}
where $\mathcal{L}_{\text{op}}$ supervises the pre-crossover population
proposal on valid teacher-selected coordinates, and
$\mathcal{L}_{\text{cand}}$ supervises the resulting candidate
transition.
We set $\lambda_{\,\text{cand}}=0.25$.
Teacher crossover and boundary-repair masks are used to
construct valid supervision targets and exclude coordinates whose values
are determined by stochastic repair. The weighting coefficient
$\lambda_{\,\text{cand}}$ and other training configurations are specified
in the experimental setup.

This stage transfers the realized search behavior demonstrated by the
teachers without embedding their handcrafted update rules into the A2M
architecture.

\subsubsection{Stage II: Differentiable Online Learning}

Trajectory imitation exposes A2M only to states generated by the teachers \cite{ross2011reduction}.
We therefore continue training on model-generated search
trajectories. Given a population $(\bm{X}_t,\bm{y}_t)$, A2M generates
and evaluates $\bm{U}_t$, after which strict one-to-one selection produces
the next population $(\bm{X}_{t+1},\bm{y}_{t+1})$.

For minimization, the online objective balances population improvement
and search diversity:
\begin{equation}
\left\{
\begin{aligned}
    \mathcal{L}_{\text{online}}
    &=
    \frac{
        \overline{y}_{t+1}-\overline{y}_{t}
    }{
        \max(|\overline{y}_{t}|,\epsilon)
    }
    -
    \frac{\lambda_{\,\text{div}}}{D}
    \!\cdot\!
    \sum_{j=1}^{D}
    \operatorname{std}_{i}\!
    \bigl(
        X_{t+1,i,j}
    \bigr),
    \\[0.5ex]
    \overline{y}_{t}
    &=
    \frac{1}{N}\!\cdot\!\sum_{i=1}^{N}y_{t,i},
\end{aligned}
\right.
\label{eq:online_loss}
\end{equation}
where $\epsilon$ is a small numerical stabilizer and
$\lambda_{\,\text{div}}$ controls the diversity regularization.
We set $\epsilon=10^{-8}$ and $\lambda_{\,\text{div}}=0.005$ in all experiments.
The first term encourages a reduction in mean population fitness, while
the second discourages premature population collapse. Losses from
different training dimensions are aggregated with equal weight.

Gradients propagate through the differentiable objective function,
signed population attention, and the straight-through crossover path,
but not through the discrete acceptance decisions. After each
generation, the selected population is detached before being appended to
the history. Stage II therefore optimizes one-generation transitions
under A2M's own state distribution rather than backpropagating through
the entire optimization trajectory.

The continuous and mixed-variable instantiations share the same
architecture and online-learning objective. They differ in the teacher
trajectories used for Stage I and in the objective interface used to
evaluate those trajectories. In the mixed-variable setting, teacher
FEs are performed through the phenotype map in
Eq.~\eqref{eq:mixed_objective}, whereas the formal Stage-II environment
remains continuous.

\subsection{Deployment as an Iterative Optimizer}
\label{sec:a2m_deployment}

After training, A2M is deployed as an iterative population-based
optimizer with fixed model parameters. Neither teacher trajectories nor
teacher-specific update rules are required during deployment. Given an
initial population $\bm{X}_0$, its true objective values $\bm{y}_0$ are
first evaluated, and the resulting state is repeated to initialize the
history window:
\begin{equation}
    \mathcal{H}_0
    =
    \!\Bigl\{
        \!
        \bigl(\bm{X}_0,\bm{y}_0\bigr),
        \ldots,
        \bigl(\bm{X}_0,\bm{y}_0\bigr)
    \Bigr\}.
\label{eq:history_initialization}
\end{equation}

At generation $t$, the history is first aligned according to the current
fitness ordering. A2M then performs one forward pass to generate the
complete candidate population, which is subsequently evaluated through
the corresponding objective interface:
\begin{equation}
\left\{
\begin{aligned}
    \bm{U}_t
    &=
    \pi_{\theta}\!
    \bigl(
        \overline{\mathcal{H}}_t
    \bigr),
    \\[0.5ex]
    \bm{y}^{U}_t
    &=
    \widetilde{f}(\bm{U}_t),
\end{aligned}
\right.
\label{eq:a2m_inference}
\end{equation}
where $\overline{\mathcal{H}}_t$ denotes the aligned history. For continuous
optimization, $\widetilde{f}=f$, whereas for mixed-variable optimization,
$\widetilde{f}=f_q\circ\,\mathcal{P}_{q,D}$.

The evaluated candidates then undergo strict one-to-one selection against
their corresponding parents. The selected population is appended to the
history and the oldest state is discarded. The process is repeated until the FE budget is exhausted.
Although the model parameters remain fixed during deployment, the
generated search behavior adapts dynamically to the evolving population
history.

For expensive optimization, evaluating all A2M candidates at every
iteration would consume the limited FE budget too quickly.
We therefore keep the trained continuous A2M fixed as a candidate
generator and couple it with an independent X-to-Y predictor that ranks
the generated candidates using observations collected on the current
test function. Only a small subset of candidates is submitted for
FEs. The predictor architecture, offline training, test-time
adaptation, and candidate-selection procedure are detailed in Section S-I of the
supplementary file.

\begin{table*}[t]
  \centering
  \caption{Summary of Average Ranks and Pairwise Comparisons across the Three Optimization Scenarios over 20 Independent Runs Per Configuration.}
  \label{tab:cross_scenario_summary}
  \begingroup
  \setlength{\tabcolsep}{3pt}
  \renewcommand{\arraystretch}{1.15}
  \footnotesize
  \resizebox{\textwidth}{!}{%
  \begin{tabular}{l|cc|cc|cc|cc|cc|cc}
    \hline
    \multirow{3}{*}{Algorithm} & \multicolumn{4}{c|}{Continuous optimization} & \multicolumn{4}{c|}{Mixed-variable optimization} & \multicolumn{4}{c}{Expensive optimization} \\
    \cline{2-13}
      & \multicolumn{2}{c|}{BBOB (120)} & \multicolumn{2}{c|}{CEC (49)} & \multicolumn{2}{c|}{MV-BBOB (120)} & \multicolumn{2}{c|}{EOPCCV (30)} & \multicolumn{2}{c|}{BBOB (120)} & \multicolumn{2}{c}{CEC (49)} \\
    \cline{2-13}
      & Avg. rank & $+/-/=$ & Avg. rank & $+/-/=$ & Avg. rank & $+/-/=$ & Avg. rank & $+/-/=$ & Avg. rank & $+/-/=$ & Avg. rank & $+/-/=$ \\
    \hline
    GTDE~\cite{9804806} & 4.60 & 12/106/2 & 2.84 & 23/26/0 & 5.70 & 14/106/0 & 6.27 & 0/30/0 & 5.28 & 0/120/0 & 5.96 & 3/46/0 \\
    sDMS-PSO~\cite{7257290} & 8.26 & 0/120/0 & 8.12 & 3/46/0 & -- & -- & -- & -- & -- & -- & -- & -- \\
    SAHLPSO~\cite{tao2021self} & 6.62 & 0/120/0 & 7.71 & 0/49/0 & -- & -- & -- & -- & -- & -- & -- & -- \\
    DE$_{\textup{MV}}$~\cite{lin2018hybrid} & -- & -- & -- & -- & 10.26 & 1/119/0 & 4.13 & 3/27/0 & -- & -- & -- & -- \\
    PSO$_{\textup{MV}}$~\cite{wang2021particle} & -- & -- & -- & -- & 9.50 & 0/120/0 & 8.73 & 0/30/0 & -- & -- & -- & -- \\
    SADE-AMSS~\cite{9971764} & -- & -- & -- & -- & -- & -- & -- & -- & 3.98 & 7/113/0 & 2.65 & 13/36/0 \\
    GP-LogEI~\cite{ament2023unexpected} & -- & -- & -- & -- & -- & -- & -- & -- & 7.15 & 0/120/0 & 4.80 & 2/47/0 \\
    EvoTF~\cite{lange2024evolution} & 5.95 & 2/116/2 & 7.39 & 6/43/0 & 7.06 & 5/115/0 & 9.30 & 0/30/0 & 10.95 & 0/120/0 & 11.00 & 0/49/0 \\
    POM~\cite{li2024pretrained} & 2.31 & 24/95/1 & 3.82 & 4/45/0 & 2.62 & 15/105/0 & 4.07 & 4/26/0 & 3.14 & 0/120/0 & 3.61 & 0/49/0 \\
    MetaDistill~\cite{han2026textttmetadistill} & 3.58 & 8/112/0 & 9.76 & 0/49/0 & 4.12 & 7/113/0 & 10.53 & 0/30/0 & 2.39 & 0/120/0 & 6.80 & 0/49/0 \\
    LLM-Generate & 9.95 & 0/120/0 & 6.94 & 3/46/0 & 10.03 & 1/119/0 & 4.70 & 1/29/0 & 8.79 & 0/120/0 & 7.65 & 0/49/0 \\
    EoH~\cite{liu2024evolution} & 5.55 & 8/110/2 & 5.49 & 12/37/0 & 3.87 & 21/99/0 & 2.80 & 4/26/0 & 8.89 & 0/120/0 & 8.82 & 0/49/0 \\
    MEoH~\cite{yao2025multi} & 7.16 & 4/114/2 & 3.86 & 14/35/0 & 6.23 & 11/109/0 & 4.83 & 5/25/0 & 7.99 & 0/120/0 & 8.96 & 0/49/0 \\
    ReEvo~\cite{ye2024reevo} & 10.50 & 0/119/1 & 7.73 & 1/48/0 & 4.85 & 17/103/0 & 9.07 & 0/30/0 & 6.37 & 0/120/0 & 4.31 & 4/45/0 \\
    \hline
    \textbf{A2M} & \textbf{1.52} & -- & \textbf{2.35} & -- & \textbf{1.77} & -- & \textbf{1.57} & -- & \textbf{1.06} & -- & \textbf{1.45} & -- \\
    \hline
  \end{tabular}}
  \par\smallskip
%  \begin{minipage}{\textwidth}
%  \footnotesize
%  Average ranks are computed separately within each scenario--suite combination; lower is better, and the best rank in each column is boldfaced.
%  The counts $+/-/=$ denote configurations on which the peer algorithm is better than, worse than, or tied with A2M, respectively, following the supplementary results.
%  Parentheses give the number of configurations; CEC pools CEC2020 and CEC2026.
%  A dash denotes an unevaluated method or a self-comparison.
%  Detailed results appear in supplementary Tables~S-V--S-X.
%  \end{minipage}
  \endgroup
\end{table*}
\section{Experiments}
\label{sec:experiments}
The experiments are organized around five empirical questions aligned with the claims of Algorithm Modelization. Q1 asks whether heterogeneous algorithm trajectories can be transformed into a standalone optimizer that surpasses its sources and generalizes beyond its training settings. Q2 tests whether the same architecture and learning framework extend to a different variable representation without scenario-specific redesign, whereas Q3 examines whether the learned search policy can be reused when FEs become scarce. Q4 isolates the contributions of heterogeneous teacher knowledge and the imitation-to-surpassing training process. Finally, Q5 investigates whether newly available community algorithms can be incorporated sequentially to strengthen an existing A2M. Thus, Q1 and Q4 evaluate the central premise and its technical realization, while Q2, Q3, and Q5 assess generality, reusability, and extensibility, respectively.

Table~\ref{tab:cross_scenario_summary} summarizes the average ranks and pairwise comparisons across the three optimization scenarios, with detailed per-configuration results in supplementary Tables~S-V--S-X. Figure~\ref{fig:cross_scenario_overview} visualizes the corresponding suite-level ranks. A2M attains the best average rank in all six scenario--suite combinations, while the strongest alternative varies across benchmarks. The following experiments therefore focus on why this advantage transfers across settings and which parts of the framework produce it.

% Summary values transcribed from supplementary Tables S-V--S-X.
% Pairwise counts retain the peer-versus-A2M direction of the source tables.

\subsection{Continuous Optimization: Learning and Surpassing Heterogeneous Optimizers}
\emph{Q1: Can trajectories from heterogeneous optimization paradigms be transformed into a standalone optimizer that improves beyond its sources and generalizes beyond the training settings?}

\begin{figure}[h]
\begin{center}
\includegraphics[width=1\linewidth]{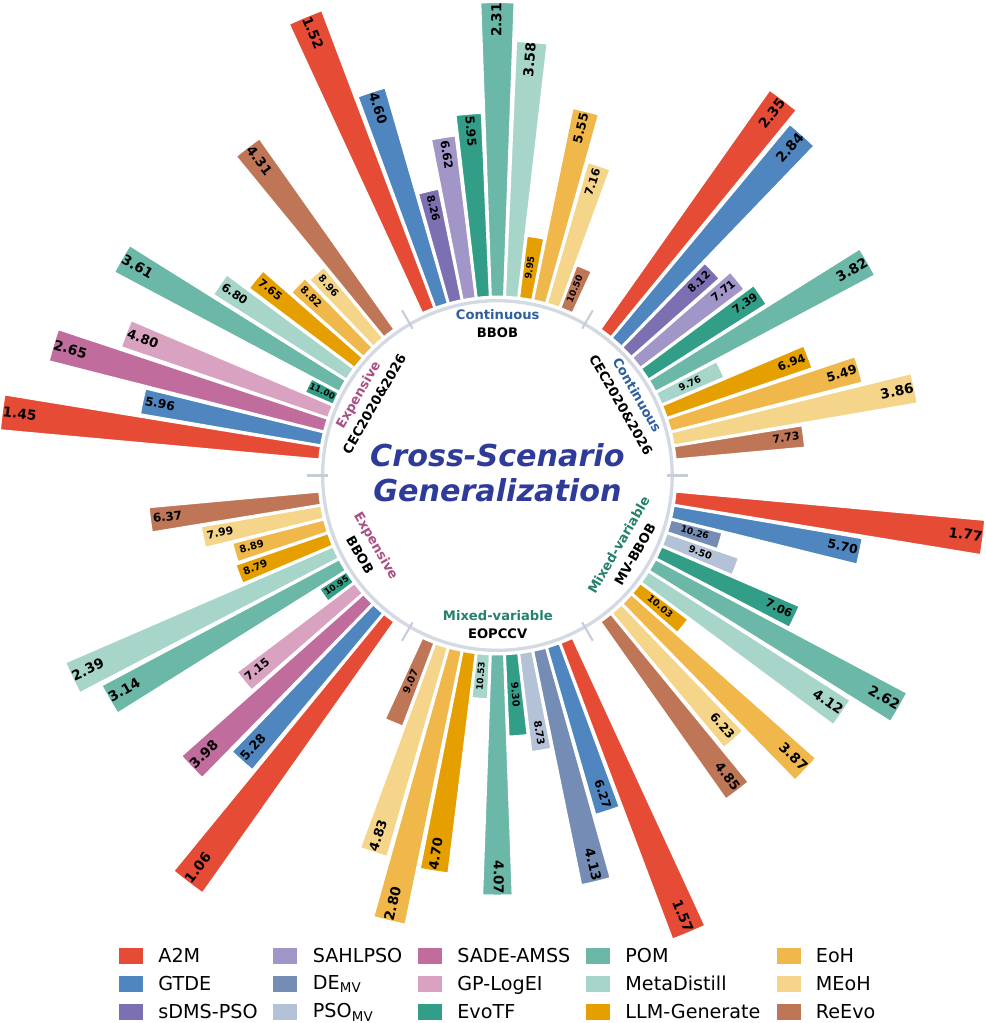}
\end{center}
  \caption{Cross-scenario comparison by average rank. The six sectors summarize continuous, mixed-variable, and expensive optimization on their respective BBOB, MV-BBOB, EOPCCV, CEC2020, and CEC2026 test suites. Numerical labels report average ranks (lower is better), while radial bar length is inversely scaled so that stronger results extend farther from the center. Only methods evaluated in a given scenario are shown.}
\label{fig:cross_scenario_overview}
\end{figure}

\subsubsection{Modelization Setting}
Continuous optimization provides the most direct test of the central premise of Algorithm Modelization. We construct the training tasks from randomly shifted instances of five functions, denoted as TF1--TF5~\cite{li2024pretrained}, over $D\in\{20,50,100,200\}$. Resampling shift vectors and initial populations exposes A2M to diverse search processes while preserving the underlying landscape structures. Detailed definitions of these functions are provided in Section~S-II of the supplementary file.

Stage I draws computational experience from three teachers with fundamentally different search mechanisms. GTDE~\cite{9804806} represents population-based evolutionary search, GP-LogEI~\cite{ament2023unexpected} represents surrogate-driven Bayesian optimization, and Projected-Lion~\cite{chen2023symbolic} contributes gradient-based refinement. Their executions are converted into the common population-transition representation of Section~\ref{sec:two_stage_learning}, allowing behaviors that originate from different computational procedures to be learned through the same trajectory interface. Section~S-III of the supplementary file details the population-based adaptations of GP-LogEI and Lion.

For every teacher and function--dimension pair, we generate 108 shifted tasks, yielding $6480$ trajectories. Each trajectory contains 100 evaluated population states, from which 615,600 history--transition windows of length five are extracted. Stage II then removes teacher supervision and trains A2M on states induced by its own search process. The resulting checkpoint is fixed throughout the continuous-optimization evaluation. Complete training settings appear in Section~S-IV of the supplementary file.

\subsubsection{Evaluation Design}
All evaluation suites are held out from training. BBOB~\cite{elhara2019coco} contains 24 functions evaluated at $D\in\{20,50,100,200,500\}$, giving 120 configurations that jointly test cross-function transfer and dimensional extrapolation. CEC2020~\cite{mohamed2020problem} contributes 10 functions at $D\in\{10,20\}$, while CEC2026~\cite{Chen2026CEC} contributes 29 functions at $D=30$, giving another 49 configurations from different benchmark families. Unless stated otherwise, each configuration uses $N=100$, 10,000 FEs, and 20 independent runs. We report the final best-so-far objective value through its mean and standard deviation, together with average ranks and pairwise win/loss/tie counts relative to A2M.

The ten baselines distinguish learning algorithmic experience from competing routes to optimizer construction. GTDE, sDMS-PSO~\cite{7257290}, and SAHLPSO~\cite{tao2021self} are conventional population-based optimizers; GTDE is additionally a teacher and therefore provides a direct test of whether A2M can improve beyond inherited behavior. EvoTF~\cite{lange2024evolution}, POM~\cite{li2024pretrained}, and MetaDistill~\cite{han2026textttmetadistill} are learned optimizers, evaluated using their released checkpoints. LLM-Generate, EoH~\cite{liu2024evolution}, MEoH~\cite{yao2025multi}, and ReEvo~\cite{ye2024reevo} instead construct optimizers through LLM generation or evolutionary refinement. These four methods use DeepSeek-V4~\cite{xu2026deepseek} as a common backbone and are prompted for the continuous setting.

\subsubsection{Results}
The continuous BBOB summary in Table~\ref{tab:cross_scenario_summary}, with per-configuration results in supplementary Table~S-V, first tests whether modelization yields an independent optimization capability rather than a reproduction of a teacher. A2M records 106/12/2 wins/losses/ties against GTDE and attains the best average rank of 1.52, compared with 4.60 for its evolutionary teacher. The advantage is not confined to the source algorithm: A2M also ranks ahead of POM (2.31) and MetaDistill (3.58), with corresponding records of 95/24/1 and 112/8/0. Heterogeneous computational experience is therefore consolidated into a standalone search policy that is stronger than both an original teacher and alternative learned optimizers.

\begin{figure*}[h]
\begin{center}
\includegraphics[width=1\linewidth]{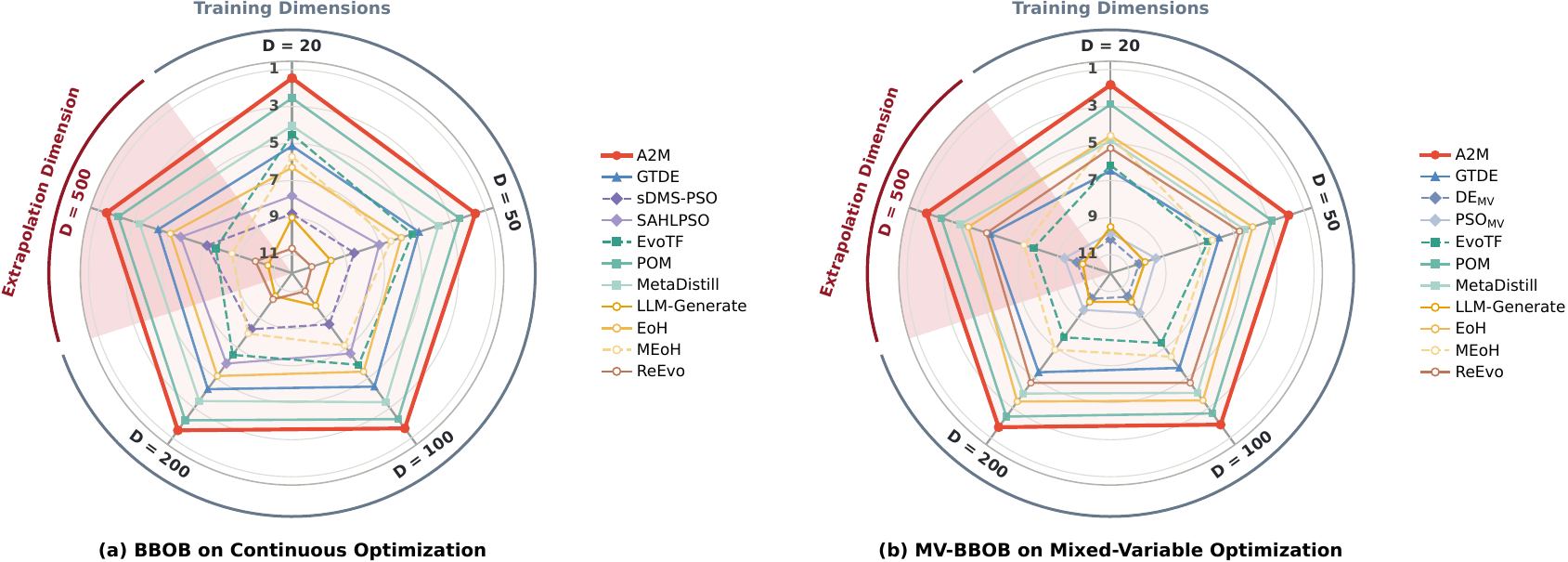}
\end{center}
  \caption{Cross-dimensional generalization on (a) BBOB (continuous optimization scenario) and (b) MV-BBOB (mixed-variable optimization scenario). Each polygon connects an algorithm's average ranks at five dimensions; lower ranks lie farther from the center. Dimensions $D\in\{20,50,100,200\}$ are used during training, whereas $D=500$ is held out for extrapolation.}
\label{fig:dimension_generalization}
\end{figure*}

Figure~\ref{fig:dimension_generalization} (a) separates dimensional generalization from the pooled BBOB result. A2M maintains the best average-rank profile over all four training dimensions and at $D=500$, which is never seen in either training stage. At this extrapolated dimension, A2M achieves the best reported mean on 16 of the 24 functions in supplementary Table~S-V without additional training or dimension-specific parameters. Moreover, although gradient-based behavior contributes to Stage I, inference requires only FEs; the transferred behavior is expressed by the common population-transition model rather than by access to the original teacher computations.

Transfer across benchmark families provides a complementary test. Table~\ref{tab:cross_scenario_summary} summarizes the 49 CEC2020 and CEC2026 configurations, while supplementary Table~S-VI reports their per-configuration results. A2M again obtains the best average rank (2.35), ahead of GTDE (2.84) and POM (3.82). It wins 45 configurations against POM and 37 against EoH, while retaining a 26/23 advantage over GTDE. Together, the BBOB and CEC results answer Q1 by showing that A2M not only improves beyond its source optimizer, but also transfers the learned search capability to unseen functions and dimensions while remaining competitive with learned and LLM-designed alternatives.

\subsection{Mixed-Variable Optimization: Generality across Variable Representations}
\emph{Q2: Can the same modelization architecture and learning framework be instantiated for a different variable representation without scenario-specific model redesign?}

\subsubsection{Instantiation without Architectural Redesign}
This experiment tests architectural generality rather than direct checkpoint transfer. We train a separate mixed-variable A2M while retaining the architecture, history representation, and two-stage learning framework used for continuous optimization. Only the teacher experience and its evaluation interface are changed. Consequently, any adaptation to the new search space must be acquired from trajectories instead of being built into a specialized network component.

Stage I collects GTDE trajectories on mixed-variable counterparts of TF1--TF5 over $D\in\{20,50,100,200\}$. The 108 shifted tasks generated for each function--dimension pair produce $2160$ trajectories and 205,200 history--transition windows. The optimizer maintains a continuous genotype, but every FE is performed after applying the phenotype map in Eq.~\eqref{eq:mixed_objective}. Its fitness responses and selection outcomes therefore encode the mixed-variable landscape while its population states retain the common representation required by A2M.

Stage II uses the same differentiable online-learning objective as the continuous instantiation. As formulated in Section~\ref{sec:two_stage_learning}, the online environment remains continuous: no rounding layer, discrete operator, or variable-type-specific module is added to the model. This design isolates whether changing algorithmic experience, rather than redesigning the architecture, is sufficient to acquire behavior for a different variable representation.

\subsubsection{Evaluation Design}
We evaluate two complementary forms of mixed-variable search. MV-BBOB combines continuous and integer variables across the 24 BBOB functions and heterogeneous search spaces~\cite{9336648}; it is tested at $D\in\{20,50,100,200,500\}$, with its construction detailed in Section~S-V of the supplementary file. EOPCCV~\cite{9464165} combines continuous and categorical variables and contains 30 problems at $D=10$. Both use 20 runs, $N=100$, and 10,000 FEs.

The baseline set tests the shared architecture against methods that obtain mixed-variable capability in different ways. $\mathrm{DE}_{\mathrm{MV}}$~\cite{lin2018hybrid} and $\mathrm{PSO}_{\mathrm{MV}}$~\cite{wang2021particle} contain dedicated mixed-variable mechanisms. GTDE~\cite{9804806}, the teacher for this instantiation, and the released EvoTF, POM, and MetaDistill checkpoints instead generate continuous proposals whose discrete components are projected to feasible values before each FE. LLM-Generate, EoH, MEoH, and ReEvo use DeepSeek-V4 to construct optimizers specifically for the mixed-variable scenario.

\subsubsection{Results}
Comparison with the specialized baselines directly tests whether mixed-variable operators must be embedded in the model. Supplementary Tables~\mbox{S-VII} and~\mbox{S-VIII} provide the per-configuration results for MV-BBOB and EOPCCV, respectively. On MV-BBOB, A2M outperforms $\mathrm{DE}_{\mathrm{MV}}$ on 119 of 120 configurations and $\mathrm{PSO}_{\mathrm{MV}}$ on all 120. On the categorically structured EOPCCV benchmark, the corresponding records are 27/3 and 30/0. Thus, the common continuous population representation remains effective for both integer and categorical variables when feasibility is imposed through the evaluation interface.

The mixed-variable summary in Table~\ref{tab:cross_scenario_summary} reinforces this conclusion across both benchmark structures. A2M ranks first on MV-BBOB and EOPCCV with average ranks of 1.77 and 1.57, respectively, while the strongest alternative changes from POM (2.62) on the continuous--integer suite to EoH (2.80) on the continuous--categorical suite. Figure~\ref{fig:dimension_generalization} (b) further shows that A2M retains the leading average-rank profile from the training dimensions to the unseen $D=500$ MV-BBOB setting. Relative to GTDE, A2M records 106/14 on MV-BBOB and 30/0 on EOPCCV, demonstrating improvement beyond the scenario-specific teacher. These results answer Q2 by showing that the same A2M architecture and modelization procedure can be retained across different variable representations, with adaptation achieved through new computational experience rather than scenario-specific redesign.

\begin{figure*}[h]
\begin{center}
\includegraphics[width=0.8\linewidth]{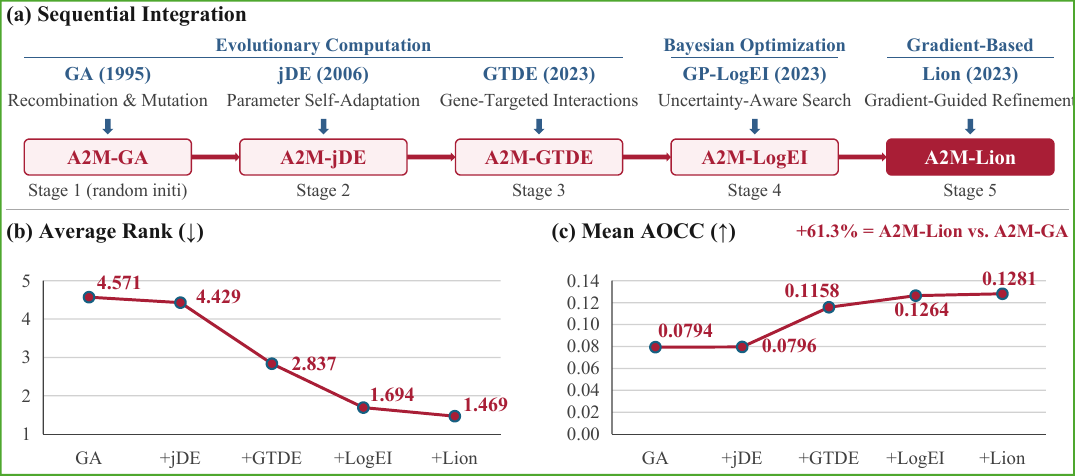}
\end{center}
  \caption{Continual evolution of A2M through sequential teacher integration. (a) A2M-GA starts from random initialization; each subsequent checkpoint inherits its predecessor and learns only from the newly introduced community algorithm. (b) Pooled average rank and (c) mean AOCC~\cite{van2025explainable} on CEC2020 and CEC2026 over 20 independent runs per configuration. Lower rank and higher AOCC are better.}
\label{fig:continual_evolution}
\end{figure*}

\begin{table*}[htbp]
  \centering
  \caption{Continual-evolution results on the CEC2020 and CEC2026 test suites.}
    \resizebox{13cm}{!}{\begin{tabular}{cc|ccccc}
    \hline
    \multicolumn{1}{c|}{} & \textit{D} & A2M-GA & A2M-jDE & A2M-GTDE & A2M-LogEI & A2M-Lion \bigstrut\\
    \hline
    \multicolumn{1}{c|}{\multirow{2}[2]{*}{CEC2020-F1}} & 10    & 1.33E+10(4.14E+09) & 1.44E+10(4.21E+09) & 3.18E+08(4.84E+08) & 3.49E+06(2.81E+06) & \textbf{3.12E+06(2.71E+06)} \bigstrut[t]\\
    \multicolumn{1}{c|}{} & 20    & 5.17E+10(1.13E+10) & 5.24E+10(1.17E+10) & 2.12E+09(1.06E+09) & \textbf{2.96E+07(2.44E+07)} & 3.28E+07(5.71E+07) \bigstrut[b]\\
    \hline
    \multicolumn{1}{c|}{\multirow{2}[2]{*}{CEC2020-F2}} & 10    & 2.71E+03(1.36E+02) & 2.67E+03(1.78E+02) & 2.24E+03(1.83E+02) & \textbf{2.13E+03(2.31E+02)} & 2.17E+03(3.09E+02) \bigstrut[t]\\
    \multicolumn{1}{c|}{} & 20    & 5.83E+03(3.37E+02) & 5.89E+03(2.39E+02) & 5.06E+03(7.33E+02) & \textbf{4.35E+03(4.37E+02)} & 4.73E+03(4.45E+02) \bigstrut[b]\\
    \hline
    \multicolumn{1}{c|}{\multirow{2}[2]{*}{CEC2020-F3}} & 10    & 1.09E+03(6.43E+01) & 1.09E+03(7.43E+01) & \textbf{7.29E+02(8.21E+00)} & 7.34E+02(7.04E+00) & 7.34E+02(4.34E+00) \bigstrut[t]\\
    \multicolumn{1}{c|}{} & 20    & 2.05E+03(1.63E+02) & 2.04E+03(1.75E+02) & 8.06E+02(2.33E+01) & 8.04E+02(1.53E+01) & \textbf{8.00E+02(1.17E+01)} \bigstrut[b]\\
    \hline
    \multicolumn{1}{c|}{\multirow{2}[2]{*}{CEC2020-F4}} & 10    & 8.83E+04(6.18E+04) & 1.02E+05(9.17E+04) & 1.91E+03(6.24E+00) & 1.90E+03(1.55E+00) & \textbf{1.90E+03(1.75E+00)} \bigstrut[t]\\
    \multicolumn{1}{c|}{} & 20    & 3.35E+06(1.59E+06) & 3.18E+06(1.57E+06) & 2.30E+03(4.95E+02) & 1.93E+03(2.22E+01) & \textbf{1.92E+03(1.16E+01)} \bigstrut[b]\\
    \hline
    \multicolumn{1}{c|}{\multirow{2}[2]{*}{CEC2020-F5}} & 10    & 1.17E+06(1.05E+06) & 9.26E+05(7.46E+05) & 2.38E+05(2.05E+05) & 2.20E+04(1.97E+04) & \textbf{1.29E+04(2.09E+04)} \bigstrut[t]\\
    \multicolumn{1}{c|}{} & 20    & 1.65E+07(7.32E+06) & 1.60E+07(5.14E+06) & 1.64E+06(1.71E+06) & \textbf{4.58E+05(2.30E+05)} & 5.73E+05(4.63E+05) \bigstrut[b]\\
    \hline
    \multicolumn{1}{c|}{\multirow{2}[2]{*}{CEC2020-F6}} & 10    & 2.16E+03(9.03E+01) & 2.11E+03(1.32E+02) & 1.88E+03(1.04E+02) & 1.79E+03(6.56E+01) & \textbf{1.78E+03(6.07E+01)} \bigstrut[t]\\
    \multicolumn{1}{c|}{} & 20    & 3.67E+03(2.36E+02) & 3.59E+03(2.08E+02) & 2.54E+03(2.04E+02) & 2.06E+03(1.43E+02) & \textbf{2.04E+03(1.61E+02)} \bigstrut[b]\\
    \hline
    \multicolumn{1}{c|}{\multirow{2}[2]{*}{CEC2020-F7}} & 10    & 1.05E+06(8.15E+05) & 1.16E+06(8.87E+05) & 1.64E+05(1.75E+05) & \textbf{7.46E+03(2.56E+03)} & 7.64E+03(1.92E+03) \bigstrut[t]\\
    \multicolumn{1}{c|}{} & 20    & 5.52E+06(3.50E+06) & 6.72E+06(3.92E+06) & 8.77E+05(1.71E+06) & 2.51E+05(1.68E+05) & \textbf{2.24E+05(1.27E+05)} \bigstrut[b]\\
    \hline
    \multicolumn{1}{c|}{\multirow{2}[2]{*}{CEC2020-F8}} & 10    & 3.19E+03(2.71E+02) & 3.13E+03(3.68E+02) & 2.32E+03(9.36E+00) & 2.31E+03(3.67E+00) & \textbf{2.31E+03(5.79E-01)} \bigstrut[t]\\
    \multicolumn{1}{c|}{} & 20    & 6.97E+03(2.86E+02) & 6.89E+03(3.10E+02) & 5.93E+03(1.28E+03) & \textbf{5.43E+03(1.60E+03)} & 5.53E+03(1.91E+03) \bigstrut[b]\\
    \hline
    \multicolumn{1}{c|}{\multirow{2}[2]{*}{CEC2020-F9}} & 10    & 2.84E+03(6.44E+01) & 2.84E+03(5.20E+01) & \textbf{2.73E+03(2.07E+01)} & 2.73E+03(2.10E+01) & 2.73E+03(2.18E+01) \bigstrut[t]\\
    \multicolumn{1}{c|}{} & 20    & 3.36E+03(1.14E+02) & 3.41E+03(1.04E+02) & 2.90E+03(3.84E+01) & \textbf{2.88E+03(3.21E+01)} & 2.90E+03(2.01E+01) \bigstrut[b]\\
    \hline
    \multicolumn{1}{c|}{\multirow{2}[2]{*}{CEC2020-F10}} & 10    & 4.13E+03(2.52E+02) & 4.02E+03(2.43E+02) & 3.09E+03(7.50E+01) & \textbf{3.01E+03(4.61E+01)} & 3.03E+03(5.50E+01) \bigstrut[t]\\
    \multicolumn{1}{c|}{} & 20    & 1.14E+04(2.33E+03) & 1.12E+04(2.13E+03) & 3.38E+03(2.48E+02) & 3.08E+03(1.23E+02) & \textbf{2.99E+03(4.47E+01)} \bigstrut[b]\\
    \hline
    \multicolumn{1}{c|}{CEC2026-F1} & 30    & 1.14E+11(1.71E+10) & 1.14E+11(1.69E+10) & 6.76E+09(2.57E+09) & 9.28E+08(5.32E+08) & \textbf{4.65E+08(3.01E+08)} \bigstrut[t]\\
    \multicolumn{1}{c|}{CEC2026-F2} & 30    & 8.94E+49(2.85E+50) & 4.64E+49(1.33E+50) & 7.43E+35(2.24E+36) & 4.79E+31(1.52E+32) & \textbf{1.33E+27(2.57E+27)}\\
    \multicolumn{1}{c|}{CEC2026-F3} & 30    & 2.99E+05(6.64E+04) & 3.12E+05(7.15E+04) & 6.91E+04(7.65E+03) & \textbf{5.83E+04(8.77E+03)} & 6.35E+04(1.15E+04)\\
    \multicolumn{1}{c|}{CEC2026-F4} & 30    & 4.31E+04(6.33E+03) & 4.29E+04(6.41E+03) & 1.07E+03(3.49E+02) & 5.82E+02(2.97E+01) & \textbf{5.68E+02(3.10E+01)}\\
    \multicolumn{1}{c|}{CEC2026-F5} & 30    & 1.15E+03(4.43E+01) & 1.15E+03(4.46E+01) & 6.68E+02(3.02E+01) & \textbf{6.65E+02(3.93E+01)} & 6.65E+02(3.57E+01)\\
    \multicolumn{1}{c|}{CEC2026-F6} & 30    & 7.27E+02(1.23E+01) & 7.24E+02(1.02E+01) & 6.44E+02(1.65E+01) & 6.17E+02(4.27E+00) & \textbf{6.15E+02(4.14E+00)}\\
    \multicolumn{1}{c|}{CEC2026-F7} & 30    & 3.24E+03(2.37E+02) & 3.23E+03(2.31E+02) & 9.70E+02(5.86E+01) & 9.22E+02(2.23E+01) & \textbf{9.04E+02(1.57E+01)}\\
    \multicolumn{1}{c|}{CEC2026-F8} & 30    & 1.36E+03(3.74E+01) & 1.36E+03(3.79E+01) & 9.49E+02(2.44E+01) & 9.43E+02(4.02E+01) & \textbf{9.39E+02(3.48E+01)}\\
    \multicolumn{1}{c|}{CEC2026-F9} & 30    & 3.32E+04(4.95E+03) & 3.30E+04(5.20E+03) & 3.27E+03(8.49E+02) & 2.13E+03(5.24E+02) & \textbf{1.65E+03(3.11E+02)}\\
    \multicolumn{1}{c|}{CEC2026-F10} & 30    & 8.73E+03(3.05E+02) & 8.64E+03(2.63E+02) & \textbf{6.08E+03(1.07E+03)} & 8.29E+03(4.79E+02) & 8.42E+03(3.65E+02)\\
    \multicolumn{1}{c|}{CEC2026-F11} & 30    & 2.86E+04(9.00E+03) & 3.00E+04(7.86E+03) & 3.78E+03(1.14E+03) & 1.72E+03(3.57E+02) & \textbf{1.72E+03(1.97E+02)}\\
    \multicolumn{1}{c|}{CEC2026-F12} & 30    & 2.25E+10(5.11E+09) & 2.16E+10(4.25E+09) & 5.39E+08(4.60E+08) & 8.60E+07(7.17E+07) & \textbf{4.96E+07(2.44E+07)}\\
    \multicolumn{1}{c|}{CEC2026-F13} & 30    & 1.68E+10(5.52E+09) & 1.67E+10(5.47E+09) & 1.41E+08(2.55E+08) & 1.65E+06(1.22E+06) & \textbf{1.63E+06(1.45E+06)}\\
    \multicolumn{1}{c|}{CEC2026-F14} & 30    & 1.02E+07(5.73E+06) & 9.98E+06(5.47E+06) & 8.72E+05(3.77E+05) & 6.17E+05(4.46E+05) & \textbf{5.45E+05(4.91E+05)}\\
    \multicolumn{1}{c|}{CEC2026-F15} & 30    & 2.20E+09(8.45E+08) & 2.43E+09(7.02E+08) & 9.64E+05(1.51E+06) & \textbf{7.99E+04(3.77E+04)} & 1.02E+05(6.34E+04)\\
    \multicolumn{1}{c|}{CEC2026-F16} & 30    & 7.00E+03(8.08E+02) & 7.17E+03(9.98E+02) & 3.33E+03(3.20E+02) & 3.04E+03(3.66E+02) & \textbf{3.04E+03(4.40E+02)}\\
    \multicolumn{1}{c|}{CEC2026-F17} & 30    & 5.59E+03(2.75E+03) & 6.10E+03(4.43E+03) & 2.65E+03(1.89E+02) & 2.22E+03(2.25E+02) & \textbf{2.15E+03(1.73E+02)}\\
    \multicolumn{1}{c|}{CEC2026-F18} & 30    & 9.34E+07(4.71E+07) & 9.12E+07(4.53E+07) & 8.18E+06(3.51E+06) & 1.63E+06(1.68E+06) & \textbf{1.49E+06(1.57E+06)}\\
    \multicolumn{1}{c|}{CEC2026-F19} & 30    & 2.57E+09(1.24E+09) & 2.27E+09(1.16E+09) & 1.59E+06(1.08E+06) & \textbf{6.24E+05(4.82E+05)} & 9.73E+05(7.76E+05)\\
    \multicolumn{1}{c|}{CEC2026-F20} & 30    & 3.31E+03(1.22E+02) & 3.33E+03(9.87E+01) & 2.70E+03(1.63E+02) & \textbf{2.54E+03(1.32E+02)} & 2.58E+03(1.12E+02)\\
    \multicolumn{1}{c|}{CEC2026-F21} & 30    & 2.90E+03(5.28E+01) & 2.90E+03(5.11E+01) & 2.46E+03(1.84E+01) & \textbf{2.45E+03(3.76E+01)} & 2.46E+03(3.31E+01)\\
    \multicolumn{1}{c|}{CEC2026-F22} & 30    & 1.03E+04(2.43E+02) & 1.03E+04(2.68E+02) & 3.69E+03(5.56E+02) & 3.95E+03(2.73E+03) & \textbf{3.24E+03(2.14E+03)}\\
    \multicolumn{1}{c|}{CEC2026-F23} & 30    & 3.65E+03(1.03E+02) & 3.67E+03(1.02E+02) & 2.98E+03(4.60E+01) & 2.91E+03(5.17E+01) & \textbf{2.89E+03(4.81E+01)}\\
    \multicolumn{1}{c|}{CEC2026-F24} & 30    & 3.83E+03(1.36E+02) & 3.86E+03(1.50E+02) & 3.05E+03(1.34E+02) & \textbf{3.03E+03(4.44E+01)} & 3.03E+03(3.71E+01)\\
    \multicolumn{1}{c|}{CEC2026-F25} & 30    & 1.60E+04(3.05E+03) & 1.60E+04(3.05E+03) & 3.19E+03(6.87E+01) & 2.98E+03(2.69E+01) & \textbf{2.97E+03(2.26E+01)}\\
    \multicolumn{1}{c|}{CEC2026-F26} & 30    & 1.47E+04(1.38E+03) & 1.46E+04(1.41E+03) & 7.54E+03(5.47E+02) & 6.03E+03(3.43E+02) & \textbf{5.79E+03(3.84E+02)}\\
    \multicolumn{1}{c|}{CEC2026-F27} & 30    & 4.24E+03(2.35E+02) & 4.08E+03(1.53E+02) & 3.57E+03(4.56E+02) & 3.31E+03(2.84E+01) & \textbf{3.30E+03(4.30E+01)}\\
    \multicolumn{1}{c|}{CEC2026-F28} & 30    & 1.10E+04(1.44E+03) & 1.11E+04(1.31E+03) & 4.12E+03(3.12E+02) & \textbf{3.41E+03(5.63E+01)} & 3.42E+03(5.29E+01)\\
    \multicolumn{1}{c|}{CEC2026-F29} & 30    & 7.41E+03(1.04E+03) & 8.05E+03(1.31E+03) & 5.20E+03(3.61E+02) & 4.61E+03(3.58E+02) & \textbf{4.39E+03(2.25E+02)} \bigstrut[b]\\
    \hline
    \multicolumn{2}{c|}{$+/-/=$} & 0/49/0 & 0/49/0 & 4/45/0 & 19/30/0 & \textbf{/} \bigstrut\\
    \hline
    \multicolumn{2}{c|}{Average Rank} & 4.57 & 4.43 & 2.84 & 1.69 & \textbf{1.47} \bigstrut\\
    \hline
    \multicolumn{2}{c|}{AOCC} & 0.0794 & 0.0796 & 0.1158 & 0.1264 &  \textbf{0.1281} \bigstrut\\
    \hline
    \end{tabular}}
  \label{tab:continual_evolution_cec}
\end{table*}
\subsection{Expensive Optimization: Reusing Search Knowledge under Limited FEs}
\emph{Q3: Can the learned continuous search policy be reused unchanged when FEs become scarce?}

\subsubsection{Frozen Search Model and Adaptive FE Allocation}
This experiment changes the deployment constraint rather than the optimization knowledge stored in A2M. The continuous checkpoint is frozen and reused as a candidate generator, preserving the behaviors acquired from evolutionary, Bayesian, and gradient-based teachers. Evaluating its entire proposed population would exhaust a small FE budget rapidly, so an independent X-to-Y predictor ranks the candidates and submits only a selected subset for FEs. Search generation and FE allocation are thereby separated: A2M determines where to search, while the predictor determines which proposals justify evaluation.

The X-to-Y predictor is pretrained on trajectories generated by GTDE, GP-LogEI, and Projected-Lion on TF1--TF5 over $D\in\{20,50,100,200,500\}$. Each sample maps population coordinates to standardized objective values, with normalization computed solely from observations in the preceding five generations. The resulting checkpoint provides a common initialization for every test problem.

Before each candidate-selection round, the predictor is adapted using only observations obtained from FEs on the current problem and RBF-interpolated auxiliary samples. Every adaptation restarts from the same pretrained predictor checkpoint, whereas A2M remains fixed throughout the run. Section~S-I of the supplementary file provides the architecture, adaptation, and candidate-selection details. This design makes the distinction central to Q3 explicit: the evaluation module adapts to the current problem, but the learned optimization policy is reused without retraining.

\subsubsection{Evaluation Design}
We retain the continuous-optimization benchmarks and dimensions while reducing the budget from 10,000 to 300 FEs. BBOB is evaluated at $D\in\{20,50,100,200,500\}$, the 10 CEC2020 functions at $D\in\{10,20\}$, and the 29 CEC2026 functions at $D=30$, yielding 120 BBOB and 49 CEC configurations. The population size remains $N=100$, and every configuration is repeated 20 times.

The baseline set is adjusted to emphasize FE-efficient search. SADE-AMSS~\cite{9971764} and GP-LogEI~\cite{ament2023unexpected} represent surrogate-assisted evolutionary optimization and Bayesian optimization, respectively, while GTDE remains a common reference across the three application scenarios. EvoTF, POM, and MetaDistill use their released checkpoints. LLM-Generate, EoH, MEoH, and ReEvo use DeepSeek-V4 to construct optimizers for the limited-FE setting. All methods receive the same 300-FE budget, placing the frozen A2M and its adaptive selector against handcrafted, learned, and automatically generated alternatives.

\subsubsection{Results}
BBOB first tests whether proposals from the frozen model remain useful when only a small fraction can be evaluated. Table~\ref{tab:cross_scenario_summary} shows an average rank of 1.06 for the combined system, ahead of MetaDistill (2.39), POM (3.14), and SADE-AMSS (3.98); supplementary Table~S-IX provides the per-configuration results. It wins all 120 configurations against every baseline except SADE-AMSS and records 113/7 against this FE-efficient competitor. Thus, the search behavior learned under the standard continuous setting remains effective when an adaptive module selectively allocates FEs.

The 49 CEC2020 and CEC2026 configurations provide a cross-suite test of the same reuse mechanism. Table~\ref{tab:cross_scenario_summary} shows that A2M again ranks first, with an average rank of 1.45 versus 2.65 for SADE-AMSS and 3.61 for POM; supplementary Table~S-X provides the per-configuration results. It wins 36 configurations against SADE-AMSS, 47 against GP-LogEI, and all 49 against POM. On CEC2026, the detailed results in supplementary Table~S-X show the best reported mean for A2M on 20 of 29 functions, with GTDE, SADE-AMSS, and ReEvo leading the remaining functions.

These results answer Q3 by showing that the learned search policy can remain fixed under a substantially reduced FE budget, while an independent adaptive module changes only how its proposals are selected for evaluation. The continuous A2M can therefore be reused without retraining the underlying search model.

\subsection{Ablation Studies: Sources of A2M's Capability}
\label{sec:ablation}

\emph{Q4: Which sources of knowledge and training stages are responsible for A2M's optimization capability?}

The preceding experiments establish what the learned optimizer can do; the ablations examine how that capability is formed. We test two hypotheses underlying the technical realization of A2M: heterogeneous teachers provide complementary computational experience, and performance-driven learning advances the model beyond trajectory imitation. Supplementary Table~S-XI reports both groups on the 24 BBOB functions at $D\in\{20,50,100,200,500\}$, totaling 120 configurations. Every variant follows the continuous-optimization protocol and uses \emph{A2M (Full)} as the common reference.

\subsubsection{Effectiveness of Teacher Diversity}
\label{sec:ablation_teacher}

We test whether this principle also applies to algorithmic experience by comparing \emph{EC-based only (GTDE)}, \emph{BO-based only (GP-LogEI)}, and \emph{Gradient-based only (Lion)} with the multi-teacher model. Each single-teacher variant receives Stage-I trajectories from only its designated teacher, whereas A2M (Full) learns from all three paradigms. The variants are A2M models rather than the original teacher algorithms; their architecture, Stage-II procedure, training functions, and total number of supervised updates are held fixed.

A2M (Full) records 115/3/2, 120/0/0, and 88/29/3 wins/losses/ties against the EC-, BO-, and gradient-based variants, respectively. The gain over each single-source variant shows that the benefit is not explained by selecting a single dominant teacher. Instead, the shared trajectory representation allows evolutionary, Bayesian, and gradient-based behaviors to contribute complementary search priors within one model. Teacher diversity therefore strengthens the learned optimizer without requiring additional architecture or a different training pipeline.

\subsubsection{Contribution of the Two Training Stages}
\label{sec:ablation_stage}

The second group separates knowledge acquisition from autonomous refinement. \emph{Imitation learning only} directly evaluates the multi-teacher Stage-I checkpoint, whereas \emph{Online learning only} applies the performance-driven objective to randomly initialized parameters. Both use the same architecture as the complete pipeline. Here, online learning denotes model training on the objective-function distribution; all parameters remain fixed during benchmark evaluation.

A2M (Full) outperforms the online-only model on all 120 configurations, demonstrating that optimization feedback alone does not replace a learned algorithmic prior. It also improves on the imitation-only checkpoint in 110 configurations, showing that reproducing teacher transitions is not the endpoint of modelization: feedback from A2M's own state distribution further develops the inherited behavior.

The two ablation groups therefore answer Q4 jointly. Heterogeneous trajectories establish a broad computational prior, and the second stage turns that prior into a stronger independent search capability, directly supporting the imitation-to-surpassing design.

\subsection{Continual Evolution through Community Algorithms}
\label{sec:continual_evolution}

\emph{Q5: Can a fixed A2M architecture progressively incorporate newly available community algorithms without replaying earlier teacher data?}

Algorithm Modelization is intended to make optimization knowledge extensible rather than leaving it fixed in a completed model. We therefore organize the final experiment as a temporal sequence: each newly introduced optimizer supplies additional computational experience to the model produced at the preceding stage. The purpose is not merely to compare five teacher choices, but to test whether A2M can accumulate search capability as the optimization community develops.

\subsubsection{Community-Evolution Protocol}
We construct an illustrative sequence of five optimization milestones:
$
\mathrm{GA} (1995)$~\cite{deb1995simulated}
$\rightarrow
\mathrm{jDE} (2006)$~\cite{brest2006self}
$\rightarrow
\mathrm{GTDE} (2023)$~\cite{9804806}
$\rightarrow
\mathrm{GP\text{-}LogEI} (2023)$~\cite{ament2023unexpected}
$\rightarrow
\mathrm{Lion} (2023)$~\cite{chen2023symbolic}.
The sequence begins within evolutionary computation and then expands to other optimization paradigms. GA introduces population-level recombination through simulated binary crossover and polynomial mutation; jDE adds online self-adaptation of differential-evolution control parameters; and GTDE contributes gene-targeted population interactions. GP-LogEI introduces uncertainty-aware stable Bayesian search, after which Lion contributes gradient-guided refinement.

The architecture remains unchanged throughout the sequence. The GA stage begins from random initialization, performs imitation, and then undergoes online learning. Every later stage initializes from the frozen checkpoint produced by its predecessor and repeats the same procedure using trajectories only from the newly introduced teacher. Each teacher contributes one supervised epoch of 81 updates; online learning uses TF1--TF5 at dimensions 20, 50, 100, and 200 with equal weights and $N=100$. Consequently, new experience is incorporated without retraining from scratch or replaying all earlier teacher trajectories.

\subsubsection{Results and Discussion}
The five successive checkpoints are evaluated on CEC2020 and CEC2026 with 10,000 FEs. Figure~\ref{fig:continual_evolution}(a) summarizes the sequential integration protocol, while Fig.~\ref{fig:continual_evolution}(b)--(c) reports pooled average rank and area over the convergence curve (AOCC)~\cite{van2025explainable}. Table~\ref{tab:continual_evolution_cec} provides the corresponding per-configuration results.

The first three milestones trace the accumulation of evolutionary-search knowledge. Advancing from the GA stage to the jDE stage changes the average rank from $4.571$ to $4.429$ and mean AOCC from $0.0794$ to $0.0796$. Incorporating GTDE then produces a larger transition: average rank decreases to $2.837$, while mean AOCC rises to $0.1158$. Because the GTDE stage begins from the jDE-trained checkpoint, its gene-targeted interactions augment an already learned evolutionary prior rather than defining an independent model.

The final two milestones test whether this accumulation extends across paradigm boundaries. Introducing Bayesian surrogate modeling reduces the average rank to $1.694$ and increases mean AOCC to $0.1264$. The subsequent gradient-based stage further improves these measures to $1.469$ and $0.1281$. Table~\ref{tab:continual_evolution_cec} shows that the final checkpoint improves on the GA, jDE, GTDE, and GP-LogEI checkpoints on 49, 49, 45, and 30 of the 49 configurations, respectively. Across the complete sequence, mean AOCC increases by $61.3\%$, from $0.0794$ to $0.1281$, while the architecture remains fixed.

This progression answers Q5 by demonstrating a practical form of continual algorithm modelization: a model trained on earlier computational experience can serve as the initialization for learning newly available algorithms, including algorithms with qualitatively different inductive biases. A2M can therefore grow with the algorithmic knowledge supplied to it, rather than requiring a new model design whenever the optimization community advances.

\section{Conclusion}
\label{sec:con}
This work introduced \emph{Algorithm Modelization}, a paradigm that transforms iterative algorithms from fixed procedures into trainable and extensible models by treating their execution trajectories as computational experience. As its first realization, A2M formulates population-based optimization as history-conditioned next-population generation. Its dimension-independent population-temporal architecture learns both population interactions and iterative dynamics, while imitation-to-surpassing training first consolidates knowledge from heterogeneous teachers and then improves the acquired search capability through optimization feedback. Once trained, A2M operates independently without invoking its source algorithms.

Experiments validate the central premise of Algorithm Modelization. A2M surpasses its evolutionary teacher and performs competitively against conventional, learned, and automatically designed optimizers on held-out benchmarks, while generalizing to the unseen dimension of $D=500$. Separately trained instances using the same framework rank first in both continuous--integer and continuous--categorical optimization, and the frozen continuous model remains effective with only 300 FEs. Ablations verify the contributions of teacher diversity and performance-driven training. Moreover, continual modelization integrates new evolutionary, Bayesian, and gradient-based algorithms without replaying previous trajectories, improving mean AOCC by $61.3\%$.

This study remains a proof of concept: continuous and mixed-variable settings currently require separate model instances, and limited-FE optimization relies on an additional evaluation module. Future work will pursue more unified models across variable representations and problem classes, stronger continual knowledge integration, and broader applications to constrained and multi-objective optimization. Overall, the results demonstrate that algorithmic knowledge can be learned, consolidated, improved, reused, and progressively extended through trainable models.

\bibliographystyle{IEEEtran}
\bibliography{myref}

\end{document}

%\linenumbers

\title{Supplementary File of ``Algorithm Modelization: Transforming Optimization Algorithms into Evolvable Intelligence''}

\maketitle\IEEEpeerreviewmaketitle
	
	\setcounter{section}{0}
	\renewcommand\thesection{S-\Roman{section}}
	
	\setcounter{table}{0}
	\renewcommand\thetable{S-\Roman{table}}
	
	\setcounter{figure}{0}
	\renewcommand\thefigure{S-\arabic{figure}}
	
	\setcounter{equation}{0}
	\renewcommand\theequation{S-\arabic{equation}}

\maketitle
\IEEEdisplaynontitleabstractindextext
\IEEEpeerreviewmaketitle

\section{A2M for Expensive Black-Box Optimization}
\label{app:expensive_a2m}

This section describes the extension of A2M to expensive
single-objective continuous optimization. The A2M model introduced in
the main paper is kept fixed and serves only as a candidate generator.
To avoid evaluating the complete candidate population, we introduce an
independent X-to-Y predictor that estimates the relative quality of
unevaluated candidates and determines which points receive true function
evaluations (FEs).

\subsection{Independent X-to-Y Predictor}

Let
\begin{equation}
    \mathcal{A}_n
    =
    \left\{
        (\bm{x}_a,y_a)
    \right\}_{a=1}^{n},
    \qquad
    y_a=f(\bm{x}_a),
\end{equation}
denote the archive of truly evaluated solutions. At optimization round
$b$, frozen A2M generates a candidate population from the current
history, while the X-to-Y predictor assigns one scalar score to each
candidate:
\begin{equation}
\begin{aligned}
    \bm{C}_b
    &=
    \pi_{\theta}
    \left(
        \mathcal{H}_b
    \right),\\
    \widehat{\bm{z}}_b
    &=
    s_{\phi_b}
    \left(
        \widetilde{\bm{C}}_b,\bm{m}_D
    \right),
    \qquad
    \nabla_{\theta}
    \mathcal{L}_{\mathrm{X2Y}}
    =0 ,
\end{aligned}
\label{eq:expensive_x2y}
\end{equation}
where $\widetilde{\bm{C}}_b$ denotes the normalized and padded candidate
coordinates and $\bm{m}_D$ is the dimension mask. The parameters
$\theta$ and $\phi$ are completely disjoint, and A2M remains frozen
throughout expensive optimization.

The predictor receives only the current candidate coordinates and the
dimension mask. It does not use historical objective values, function or
benchmark identities, generation indices, or hidden states from A2M.
Its input encoder combines two complementary representations. A masked
vector path preserves coordinate-slot information after padding to the
maximum supported dimension, while a shared-coordinate path applies the
same coordinate encoder to every active dimension and aggregates the
resulting features symmetrically. The two representations are fused and
processed by a population Transformer, followed by a row-wise scalar
prediction head. No row positional embeddings are used, preserving
permutation equivariance with respect to candidate ordering.

\subsection{Offline Pretraining and Test-Time Adaptation}

The X-to-Y predictor is first pretrained on teacher trajectories.
For each population, the input consists only of its current coordinates,
while the regression targets are standardized objective values.
The normalization statistics are computed from preceding generations,
such that the prediction target does not introduce future information.
The pretraining objective combines point-wise regression with relative
ranking supervision:
\begin{equation}
    \mathcal{L}_{\mathrm{pre}}
    =
    \mathcal{L}_{\mathrm{reg}}
    +
    0.1\,\mathcal{L}_{\mathrm{rank}}
    +
    0.5\,\mathcal{L}_{\mathrm{elite}},
\label{eq:x2y_pretrain}
\end{equation}
where $\mathcal{L}_{\mathrm{reg}}$ is a Smooth-L1 regression loss,
$\mathcal{L}_{\mathrm{rank}}$ preserves pairwise ordering, and
$\mathcal{L}_{\mathrm{elite}}$ emphasizes the ranking of the five
best solutions in each population.

During expensive optimization, the predictor is adapted separately to
each test function using only the observations currently available in
$\mathcal{A}_n$. To augment the limited real observations, we construct
interpolation samples between nearby evaluated points and assign them
pseudo-labels using a thin-plate RBF fitted to the current archive. These
samples are used only as auxiliary supervision; they neither enter the
candidate pool nor serve as acquisition scores.

Importantly, adaptation is restarted from the same pretrained checkpoint
at every optimization round:
\begin{equation}
    \phi_b
    =
    \operatorname{Adapt}
    \left(
        \phi_0;
        \mathcal{A}_{n_b},
        \mathcal{A}^{\mathrm{aug}}_b
    \right),
    \qquad
    \phi_b
    \not\leftarrow
    \phi_{b-1}.
\label{eq:x2y_restart}
\end{equation}
The test-time objective retains the three real-data terms in
Eq.~\eqref{eq:x2y_pretrain} and adds a down-weighted regression loss on
the RBF-labeled interpolation samples:
\begin{equation}
    \mathcal{L}^{(b)}_{\mathrm{test}}
    =
    \mathcal{L}^{\mathrm{real}}_{\mathrm{reg}}
    +
    0.1\,\mathcal{L}^{\mathrm{real}}_{\mathrm{rank}}
    +
    0.5\,\mathcal{L}^{\mathrm{real}}_{\mathrm{elite}}
    +
    0.1\,\mathcal{L}^{\mathrm{aug}}_{\mathrm{reg}} .
\label{eq:x2y_adaptation_loss}
\end{equation}
In the formal configuration, each round generates 50 interpolation
samples and performs 50 AdamW adaptation steps. Only part of the X-to-Y
predictor is updated, while A2M itself is always fixed.

\subsection{Surrogate-Assisted Expensive Optimization}

The formal protocol uses a total budget of $B=300$ FEs.
It starts from $N_0=100$ evaluated solutions and subsequently
performs 40 rounds with five FEs per round.

At the beginning of round $b$, the archive is sorted by true objective
value and its best 100 solutions form the current population
$\bm{P}_b$. Since expensive optimization does not maintain a complete
A2M rollout, this evaluated population is repeated to initialize all
$K$ history positions:
\begin{equation}
    \mathcal{H}_b
    =
    \left\{
        (\bm{P}_b,\bm{z}^{P}_b),
        \ldots,
        (\bm{P}_b,\bm{z}^{P}_b)
    \right\},
    \qquad K\ \text{copies},
\label{eq:expensive_history}
\end{equation}
where $\bm{z}^{P}_b$ contains objective values standardized using the
current true archive.

A2M is then invoked once to generate $N=100$ candidates without
additional FEs. The adapted X-to-Y predictor scores these
candidates, after which previously evaluated and duplicate solutions are
removed. The remaining candidates are ranked in ascending order of their
predicted scores.

We select five candidates by combining exploitation and diversity.
The three candidates with the lowest predicted scores are selected
directly. The remaining two are chosen sequentially from the predicted
top 30 using a maximin criterion. For an unselected candidate
$\bm{c}_{b,i}$ and the current selected set $\mathcal{Q}$, its diversity
score is
\begin{equation}
    d_b(i\mid\mathcal{Q})
    =
    \min_{
        \bm{r}\in
        \mathcal{A}_{n_b}^{X}
        \cup
        \{\bm{c}_{b,j}:j\in\mathcal{Q}\}
    }
    \left\|
        \widetilde{\bm{c}}_{b,i}
        -
        \widetilde{\bm{r}}
    \right\|_2 ,
\label{eq:expensive_diversity}
\end{equation}
where distances are computed in the normalized decision space. At each
diversity step, the candidate maximizing
Eq.~\eqref{eq:expensive_diversity} is added to $\mathcal{Q}$.

Only the resulting five candidates receive FEs and are appended to the
archive. Importantly, surrogate predictions
are never written back as true fitness values. The next round is rebuilt
from the best 100 truly evaluated solutions in the updated archive,
rather than from surrogate-based one-to-one selection. Hence, the
expensive optimization process can be summarized as
\begin{equation}
\begin{aligned}
    100\ \text{initial FEs}
    \;+\;
    40\times
    \left(
        3\ \text{exploitation FEs}
        +
        2\ \text{diversity FEs}
    \right)
    =
    300\ \text{FEs}.
\end{aligned}
\end{equation}
The final solution is the best point in the true archive after all
$300$ FEs have been consumed.

\clearpage
\newpage

\section{Training Functions}
\label{app:training_functions}

\paragraph{Function suite}
Following the construction used by POM, we generate the A2M training
tasks from five continuous base functions, denoted by TF1--TF5. These
functions are deliberately separated from the benchmark instances used
for formal evaluation. Together, they form a compact and controlled set
of landscape structures, covering nonsmooth and smooth objectives,
separable and coupled variables, additive and worst-coordinate signals,
and the narrow curved valley of the Rosenbrock function. Randomly shifted
instances of this suite provide diverse search dynamics for both
teacher-trajectory collection and online training without exposing A2M
to the evaluation benchmarks.

For a decision vector $\bm{x}=(x_1,\ldots,x_D)$, each instance is
parameterized by a shift vector
$\bm{\omega}=(\omega_1,\ldots,\omega_D)$. We introduce the translated
coordinates
\begin{equation}
    z_i = x_i-\omega_i,
    \qquad i=1,\ldots,D,
    \label{eq:training_shift}
\end{equation}
Varying $\bm{\omega}$ relocates the landscape while preserving its
geometry and variable-interaction structure. The five base functions are
defined below.

\paragraph{TF1}
TF1 is the coordinate-wise absolute-value function
\begin{equation}
    f_{\mathrm{TF1}}(\bm{x};\bm{\omega})
    =
    \sum_{i=1}^{D}|z_i|.
    \label{eq:tf1}
\end{equation}
with $x_i\in[-10,10]$ and $\omega_i\in[-10,10]$. Its nonsmooth,
convex, and separable landscape provides a basic setting for learning
directed convergence toward a shifted optimum.

\paragraph{TF2}
TF2 augments the coordinate-wise absolute values with pairwise terms
that couple adjacent variables:
\begin{equation}
\begin{aligned}
    f_{\mathrm{TF2}}(\bm{x};\bm{\omega})
    =
    &\sum_{i=1}^{D-1}
    \left|z_i+z_{i+1}\right|
    +
    \sum_{i=1}^{D}|z_i|.
\end{aligned}
\label{eq:tf2}
\end{equation}
We again use $x_i\in[-10,10]$ and $\omega_i\in[-10,10]$. Compared
with TF1, the terms $|z_i+z_{i+1}|$ make TF2 non-separable and require
the optimizer to account for dependencies between neighboring
coordinates.

\paragraph{TF3}
TF3 is the shifted quadratic function
\begin{equation}
    f_{\mathrm{TF3}}(\bm{x};\bm{\omega})
    =
    \sum_{i=1}^{D}z_i^2.
    \label{eq:tf3}
\end{equation}
with $x_i\in[-100,100]$ and $\omega_i\in[-50,50]$. It provides a
smooth, strictly convex, and separable landscape in which every
coordinate contributes continuously to the objective value.

\paragraph{TF4}
TF4 measures the largest absolute coordinate deviation from the shifted
optimum:
\begin{equation}
    f_{\mathrm{TF4}}(\bm{x};\bm{\omega})
    =
    \max_{1\leq i\leq D}|z_i|.
    \label{eq:tf4}
\end{equation}
We use $x_i\in[-100,100]$ and $\omega_i\in[-50,50]$. Unlike the
preceding additive objectives, TF4 is governed entirely by the
worst-aligned coordinate. It therefore provides a sparse improvement
signal that emphasizes correction of the largest residual error.

\paragraph{TF5}
TF5 is the shifted Rosenbrock function
\begin{equation}
\begin{aligned}
    f_{\mathrm{TF5}}(\bm{x};\bm{\omega})
    =
    \sum_{i=1}^{D-1}
    \Big[
        100\left(z_i^2-z_{i+1}\right)^2
        +
        \left(z_i-1\right)^2
    \Big].
\end{aligned}
\label{eq:tf5}
\end{equation}
with $x_i\in[-100,100]$ and $\omega_i\in[-50,50]$. Its strong
variable interactions, ill conditioning, and narrow curved valley make
it substantially more difficult than TF1--TF4 and provide a challenging
setting for learning coordinated population transitions.

\begin{table}[htbp]
\centering
\caption{Summary of the training functions TF1--TF5 adopted from
POM~\cite{li2024pretrained}.}
\label{tab:training_functions}
\small
\begin{tabular}{c c c p{0.36\columnwidth}}
\hline
Function & Search range & Shift range & Landscape characteristics \\
\hline
TF1 &
$[-10,10]^D$ &
$[-10,10]^D$ &
Unimodal, separable, nonsmooth
\\

TF2 &
$[-10,10]^D$ &
$[-10,10]^D$ &
Unimodal, non-separable, adjacent-variable coupling
\\

TF3 &
$[-100,100]^D$ &
$[-50,50]^D$ &
Unimodal, separable, smooth
\\

TF4 &
$[-100,100]^D$ &
$[-50,50]^D$ &
Unimodal, non-additive, nonsmooth
\\

TF5 &
$[-100,100]^D$ &
$[-50,50]^D$ &
Non-separable, narrow-valley, ill-conditioned
\\
\hline
\end{tabular}
\end{table}

\paragraph{Task construction for A2M}
Each training task is specified by a base function, a decision
dimension, a shift vector, and an initial population. Independently
resampling $\bm{\omega}$ and the initial population produces tasks that
share the same landscape family but differ in location and initial
search state. This controlled randomization increases trajectory
diversity without requiring a large collection of unrelated handcrafted
functions.

For both the continuous and mixed-variable A2M instantiations, TF1--TF5
are constructed at $D\in\{20,50,100,200\}$. We generate 108 independently
shifted tasks for every function--dimension pair. In the continuous
setting, each task is executed by heterogeneous teacher optimizers from
different search paradigms. In the mixed-variable setting, FEs
use the phenotype mapping defined in the main paper, so the resulting
fitness values and selection outcomes encode the corresponding
mixed-variable landscape while retaining a common continuous genotype.

The online stage uses the same five base functions but continually
resamples shifts and initial populations, exposing A2M to new search
states rather than replaying a fixed set of trajectories. TF1--TF5 thus
serve as a compact training basis spanning landscape structures,
dimensions, locations, and population states; they are not intended to
approximate individual test functions. Formal evaluation remains fully
separate and is conducted on BBOB, CEC2020, CEC2026, and their
scenario-specific variants.

\clearpage
\newpage

\section{Population-Based Adaptations of GP-LogEI and Lion}
\label{app:teacher_adaptations}

To make Bayesian and gradient-based search compatible with A2M's trajectory
interface, we implement GP-LogEI and Lion as fixed-size population
optimizers. Both maintain $N$ evaluated solutions and record successive
populations together with their objective values in the same format as GTDE.

\noindent\textbf{GP-LogEI.}
We embed GP-based LogEI selection in a real-coded GA. At each generation,
a Mat\'ern-$5/2$ GP is fitted to at most the best $3N$ distinct archived samples
with finite objective values, using normalized decision coordinates.
Tournament selection, simulated binary crossover, and polynomial mutation
generate $10N$ offspring. LogEI ranks these offspring together with the $N$
parents, and the top $N$ candidates admissible under the remaining FE budget
form the next population. Only previously unseen selected points require FEs;
cached objective values are reused. To stabilize trajectory collection, we
combine robust target scaling, training-point deduplication, and adaptive
Cholesky jitter, with distance-based novelty ranking as a fallback when a GP
cannot be fitted.

\smallskip
\noindent\textbf{Lion.}
We extend Lion to a population by assigning each solution its own momentum
vector. Updates are performed in normalized coordinates and projected onto
$[0,1]^D$, making the step size relative to each variable's search range.
Each parent generates one offspring using the sign of a momentum--gradient
mixture. After FEs, the best $N$ solutions from the combined parent--offspring
pool survive, with their updated momentum states carried forward.
Per-solution RMS gradient normalization, gradient clipping, and cosine
step-size decay stabilize the updates; weight decay is omitted to avoid
biasing the search toward an arbitrary origin. Gradients are obtained by
automatic differentiation of a white-box implementation of the objective,
and gradient calls are tracked separately from FEs. This access is used only
to generate teacher trajectories: the deployed A2M requires objective values
but no gradients.

\clearpage
\newpage
\section{Model Architecture and Hyperparameters}
\label{app:model_hyperparameters}

\subsection{A2M Transformer Architecture}
\label{app:history_pom_architecture}

The continuous and mixed-variable optimizers use the same A2M transformer
architecture. They differ only in the supervision data and in the mapping from
the continuous optimizer state to the phenotype evaluated by the objective
function.

\subsubsection{Population History and Alignment}
\label{app:population_history}

At generation $t$, the optimizer maintains a population
\begin{equation}
  \mathcal{P}_t
  =
  \left\{
  \left(\mathbf{x}_{t,i},y_{t,i}\right)
  \right\}_{i=1}^{N},
  \qquad
  y_{t,i}=f(\mathbf{x}_{t,i}),
\end{equation}
where $N=100$. The model receives the most recent $K=5$ population states:
\begin{equation}
  \mathcal{H}_t
  =
  \left\{
  \left(\mathbf{X}_{t-k},\mathbf{y}_{t-k}\right)
  \right\}_{k=0}^{K-1},
\end{equation}
where
$\mathbf{X}_{t-k}\in\mathbb{R}^{N\times D}$ and
$\mathbf{y}_{t-k}\in\mathbb{R}^{N}$.

At the first generation, the initial population and its fitness values are
repeated $K$ times to construct a complete history. After each generation, the
oldest population state is removed and the newly selected population is
appended.

Because one-to-one selection preserves individual identity, all historical
populations are aligned using the fitness order of the current population.
Let
\begin{equation}
  \pi_t
  =
  \operatorname{argsort}\left(\mathbf{y}_t\right)
\end{equation}
denote the current best-to-worst permutation. The aligned history is
\begin{equation}
  \overline{\mathbf{X}}_{\tau,i}
  =
  \mathbf{X}_{\tau,\pi_t(i)},
  \qquad
  \overline{y}_{\tau,i}
  =
  y_{\tau,\pi_t(i)},
  \qquad
  \tau\in\{t-K+1,\ldots,t\}.
  \label{eq:history_alignment}
\end{equation}
Thus, row $i$ represents the same individual throughout the historical
window, while row $1$ corresponds to the currently best individual.

\subsubsection{Fitness Features}
\label{app:fitness_features}

Each individual is represented by two fitness features. The first is the
population-standardized fitness
\begin{equation}
  z_{\tau,i}
  =
  \frac{
    \overline{y}_{\tau,i}
    -
    \operatorname{mean}_j
    \left(\overline{y}_{\tau,j}\right)
  }{
    \max\left(
      \operatorname{std}_j
      \left(\overline{y}_{\tau,j}\right),
      \epsilon
    \right)
  },
  \qquad
  \epsilon=10^{-8}.
\end{equation}
The second is the standardized actual rank
\begin{equation}
  r_{\tau,i}
  =
  \frac{
    \operatorname{rank}
    \left(\overline{y}_{\tau,i}\right)
    -
    \mu_{\mathrm{rank}}
  }{
    \sigma_{\mathrm{rank}}
  },
\end{equation}
where $\mu_{\mathrm{rank}}$ and $\sigma_{\mathrm{rank}}$ are computed from
the integer ranks $\{0,\ldots,N-1\}$.

The two features are embedded as
\begin{equation}
  \mathbf{h}^{y}_{\tau,i}
  =
  \operatorname{LN}
  \left(
    \operatorname{GELU}
    \left(
      \mathbf{W}_{y}
      [z_{\tau,i},r_{\tau,i}]
      +
      \mathbf{b}_{y}
    \right)
  \right)
  \in\mathbb{R}^{256}.
  \label{eq:fitness_embedding}
\end{equation}

\subsubsection{Dimension-Independent Coordinate-Set Encoding}
\label{app:coordinate_set_encoding}

The decision variables are encoded as an unordered coordinate set. Unlike the
previous fixed-coordinate representation, the final model contains no input
layer whose number of parameters depends on $D$.

For each historical population, one shared RMS value is computed over all
valid individuals and coordinates:
\begin{equation}
  s_{\tau}
  =
  \sqrt{
    \frac{1}{ND}
    \sum_{i=1}^{N}
    \sum_{d=1}^{D}
    \overline{x}_{\tau,i,d}^{\,2}
  }.
\end{equation}
The normalized coordinates are
\begin{equation}
  \widehat{x}_{\tau,i,d}
  =
  \frac{
    \overline{x}_{\tau,i,d}
  }{
    \max(s_{\tau},10^{-8})
  }.
\end{equation}
For each coordinate, the model constructs a three-dimensional relational
token
\begin{equation}
  \mathbf{c}_{\tau,i,d}
  =
  \operatorname{clip}
  \left(
  \begin{bmatrix}
    \widehat{x}_{\tau,i,d}\\
    \widehat{x}_{\tau,i,d}
    -
    \frac{1}{N}
    \sum_{j=1}^{N}\widehat{x}_{\tau,j,d}\\
    \widehat{x}_{\tau,i,d}
    -
    \widehat{x}_{\tau,1,d}
  \end{bmatrix},
  -8,8
  \right).
  \label{eq:coordinate_token}
\end{equation}
The three terms represent the normalized coordinate value, its displacement
from the population mean, and its displacement from the trajectory of the
currently best individual.

The same coordinate encoder is applied independently to every coordinate:
\begin{equation}
  \mathbf{e}_{\tau,i,d}
  =
  \operatorname{GELU}
  \left(
    \mathbf{W}_{2}
    \operatorname{GELU}
    \left(
      \mathbf{W}_{1}\mathbf{c}_{\tau,i,d}
      +
      \mathbf{b}_{1}
    \right)
    +
    \mathbf{b}_{2}
  \right),
\end{equation}
where both hidden layers have width 32. The encoded coordinates are aggregated
using symmetric mean and RMS pooling:
\begin{align}
  \mathbf{m}_{\tau,i}
  &=
  \frac{1}{D}
  \sum_{d=1}^{D}
  \mathbf{e}_{\tau,i,d},\\
  \mathbf{q}_{\tau,i}
  &=
  \sqrt{
    \frac{1}{D}
    \sum_{d=1}^{D}
    \mathbf{e}_{\tau,i,d}^{\,2}
  }.
\end{align}
The coordinate-set representation is then
\begin{equation}
  \mathbf{h}^{x}_{\tau,i}
  =
  \operatorname{LN}
  \left(
    \operatorname{GELU}
    \left(
      \mathbf{W}_{x}
      [\mathbf{m}_{\tau,i};\mathbf{q}_{\tau,i}]
      +
      \mathbf{b}_{x}
    \right)
  \right)
  \in\mathbb{R}^{256}.
  \label{eq:coordinate_set_embedding}
\end{equation}
Because the coordinate encoder is shared and the pooling operations are
symmetric, the representation is independent of coordinate ordering and can
be evaluated at dimensions not used during training.

\subsubsection{Feature Fusion}
\label{app:feature_fusion}

A learned population-position embedding
$\mathbf{e}^{\mathrm{pos}}_i\in\mathbb{R}^{256}$ is added after the current
fitness alignment. The fitness, coordinate-set, and position features are
combined by a dynamic per-token, per-channel gate:
\begin{align}
  \mathbf{u}_{\tau,i}
  &=
  [
    \mathbf{h}^{y}_{\tau,i};
    \mathbf{h}^{x}_{\tau,i};
    \mathbf{e}^{\mathrm{pos}}_i
  ],\\
  \mathbf{g}_{\tau,i}
  &=
  \operatorname{sigmoid}
  \left(
    \mathbf{W}_{g}\mathbf{u}_{\tau,i}
    +
    \mathbf{b}_{g}
  \right),\\
  \mathbf{h}^{(0)}_{\tau,i}
  &=
  \operatorname{LN}
  \left(
    \operatorname{GELU}
    \left(
      \mathbf{W}_{0}
      [
        \mathbf{h}^{y}_{\tau,i};
        \mathbf{g}_{\tau,i}\odot\mathbf{h}^{x}_{\tau,i};
        \mathbf{e}^{\mathrm{pos}}_i
      ]
      +
      \mathbf{b}_{0}
    \right)
  \right).
  \label{eq:dynamic_feature_fusion}
\end{align}
The gate is initialized to $0.5$ in every channel by setting
$\mathbf{W}_{g}=0$ and $\mathbf{b}_{g}=0$. Therefore, coordinate information
contributes from the beginning of training rather than being initialized as a
near-zero bypass.

\subsubsection{Axial History Transformer}
\label{app:axial_history_transformer}

The fused tensor has shape
$K\times N\times d_{\mathrm{model}}$, where
$K=5$, $N=100$, and $d_{\mathrm{model}}=256$.
The encoder contains four axial Transformer blocks. Each block applies:

\begin{itemize}
  \item population self-attention across the $N$ individuals at each
  historical generation;
  \item temporal self-attention across the $K$ generations for each aligned
  individual; and
  \item a position-wise feed-forward network of width 1024.
\end{itemize}

Temporal attention uses rotary position embeddings with positions
$\{-4,-3,-2,-1,0\}$ and base $10{,}000$. Population attention does not use
rotary positions. All sublayers use pre-normalization, GELU activation,
LayerScale initialized to $0.1$, and residual connections. Dropout is disabled.

The future representation is initialized using the encoded current population.
Three decoder blocks then predict the state at temporal position $+1$. Each
decoder block first performs temporal cross-attention from the future query to
the five historical states, followed by population self-attention and a
feed-forward network. Both encoder and decoder attention use four heads with
64 dimensions per head.

\subsubsection{Signed Population Attention}
\label{app:signed_operator}

The future population representations are translated into candidate proposals
by attending to the current population. For each operator head $h$, the future
representation of individual $i$ produces a query, while the current
representation of individual $j$ produces a key:
\begin{align}
  \mathbf{q}_{t,i,h}
  &=
  \tanh
  \left(
    \mathbf{W}^{Q}_{h}
    \mathbf{h}^{\mathrm{future}}_{t,i}
    +
    \mathbf{b}^{Q}_{h}
  \right),\\
  \mathbf{k}_{t,j,h}
  &=
  \tanh
  \left(
    \mathbf{W}^{K}_{h}
    \mathbf{h}^{\mathrm{current}}_{t,j}
    +
    \mathbf{b}^{K}_{h}
  \right).
\end{align}

Their interaction determines a signed population-attention coefficient
\begin{equation}
  \omega_{t,i,j}
  =
  b_{t,i,j}
  \tanh
  \left(
    \sum_{h=1}^{H_{\mathrm{op}}}
    \alpha_h
    \frac{
      \mathbf{q}_{t,i,h}^{\top}
      \mathbf{k}_{t,j,h}
    }{
      \sqrt{d_h}
    }
  \right),
  \qquad
  b_{t,i,i}=1,
  \label{eq:signed_operator}
\end{equation}
where $H_{\mathrm{op}}=4$, $d_h=64$, and $\alpha_h$ are learnable
head-mixing coefficients. The variable $b_{t,i,j}$ denotes a stochastic
structural mask that regularizes off-diagonal population interactions while
keeping each self-interaction active.

Unlike conventional softmax attention, the coefficients
$\omega_{t,i,j}$ may be positive or negative and are not constrained to sum
to one. Consequently, the model can represent both attractive and repulsive
relations among population members.

The proposal for individual $i$ is constructed directly from the aligned
current population:
\begin{equation}
  \widetilde{\mathbf{x}}_{t,i}
  =
  \sum_{j=1}^{N}
  \omega_{t,i,j}
  \overline{\mathbf{x}}_{t,j}.
  \label{eq:population_mutation}
\end{equation}
This population-level construction allows each proposal to combine information
from multiple current individuals according to the search state inferred from
the recent optimization history. The resulting proposal is subsequently
combined with its corresponding parent by the learned coordinate-level
crossover mechanism.

\subsubsection{Learned Crossover and Greedy Selection}
\label{app:crossover_selection}

For each individual, the crossover head receives three features: standardized
parent fitness, standardized current rank, and the population-standardized
cosine similarity between the parent and mutation vectors. It predicts a
scalar crossover score
\begin{equation}
  c_{t,i}
  =
  \operatorname{sigmoid}
  \left(
    \mathbf{w}_{2}^{\top}
    \operatorname{LN}
    \left(
      \operatorname{ReLU}
      \left(
        \mathbf{W}_{1}\mathbf{a}_{t,i}
        +
        \mathbf{b}_{1}
      \right)
    \right)
    +
    b_{2}
  \right),
\end{equation}
where $\mathbf{a}_{t,i}\in\mathbb{R}^{3}$ and the hidden width is four.

For each coordinate, the crossover score and an independently sampled uniform
threshold form the two logits of a hard Gumbel--Softmax decision:
\begin{equation}
  \left(
    C^{\mathrm{parent}}_{t,i,d},
    C^{\mathrm{mutation}}_{t,i,d}
  \right)
  =
  \operatorname{HardGumbelSoftmax}
  \left(
    [c_{t,i},u_{t,i,d}],
    \tau=1
  \right),
  \qquad
  u_{t,i,d}\sim\mathcal{U}(0,1).
\end{equation}
The trial individual is
\begin{equation}
  \mathbf{u}_{t,i}
  =
  \mathbf{C}^{\mathrm{parent}}_{t,i}
  \odot
  \mathbf{x}_{t,i}
  +
  \mathbf{C}^{\mathrm{mutation}}_{t,i}
  \odot
  \mathbf{v}_{t,i}.
  \label{eq:learned_crossover}
\end{equation}

Selection is fixed and contains no learned parameters. Each trial individual
competes only against the parent at the same population row:
\begin{equation}
  \mathbf{x}_{t+1,i}
  =
  \begin{cases}
    \mathbf{u}_{t,i},
    & f(\mathbf{u}_{t,i})<f(\mathbf{x}_{t,i}),\\
    \mathbf{x}_{t,i},
    & \text{otherwise}.
  \end{cases}
  \label{eq:one_to_one_selection}
\end{equation}
This one-to-one greedy rule preserves row identity and enables the temporal
alignment in Equation~\eqref{eq:history_alignment}.

\subsection{Continuous and Mixed-Variable Instantiations}
\label{app:continuous_mixed_instantiations}

\subsubsection{Continuous Optimization}
\label{app:continuous_optimization}

In the continuous setting, the optimizer state and the evaluated phenotype are
identical after bound repair:
\begin{equation}
  \boldsymbol{\theta}_{t,i}
  =
  \operatorname{Repair}
  \left(
    \mathbf{u}_{t,i}
  \right).
\end{equation}
The repaired continuous vector is evaluated by the objective function and is
also stored in the next population history.

Continuous supervised pretraining uses trajectories from three complementary
teachers: GTDE, GP-LogEI-GA, and Projected-Lion. The trajectories are
generated on randomized shifted POM TF1--TF5 tasks at dimensions
$D\in\{20,50,100,200\}$. Task identifiers 0--99 form the training set and
100--107 form the validation set. One deterministic repeat, R0, is generated
for each teacher and task.
The population-based adaptations of GP-LogEI and Lion are described in
Section~\ref{app:teacher_adaptations}.

\subsubsection{Mixed-Variable Optimization}
\label{app:mixed_variable_optimization}

The mixed-variable setting considered in this work contains continuous and
integer decision variables. The neural optimizer still maintains a fully
continuous population. Integer constraints are applied only to the phenotype
immediately before each FE.

For function $f$, let $\mathcal{I}_f$ denote its deterministic set of integer
coordinates. With integer step size one, the phenotype mapping is
\begin{equation}
  \theta_{t,i,d}
  =
  \begin{cases}
    \operatorname{round}
    \left(
      \operatorname{Repair}(u_{t,i,d})
    \right),
    & d\in\mathcal{I}_f,\\
    \operatorname{Repair}(u_{t,i,d}),
    & d\notin\mathcal{I}_f.
  \end{cases}
  \label{eq:mixed_variable_mapping}
\end{equation}
The number of integer coordinates is
\begin{equation}
  |\mathcal{I}_f|
  =
  \operatorname{clip}
  \left(
    \operatorname{round}(\rho_f D),
    1,
    D-1
  \right),
\end{equation}
where $\rho_f$ is the predefined function-specific integer ratio. For the
TF1--TF5 training functions, the ratios are
\begin{equation}
  (\rho_1,\rho_2,\rho_3,\rho_4,\rho_5)
  =
  (0.70,0.50,0.40,0.40,0.70).
\end{equation}
The integer coordinates are selected deterministically and distributed across
the decision vector.

Mixed-variable supervised pretraining uses GTDE trajectories generated on
the corresponding mixed-integer TF1--TF5 tasks. The dimensions are again
$D\in\{20,50,100,200\}$, task identifiers 0--107 are used, and each task is
generated once with repeat R0.

\subsection{Training Objectives}
\label{app:training_objectives}

\subsubsection{Supervised Pretraining}
\label{app:supervised_pretraining}

For each historical window, the teacher provides the pre-selection mutation
displacement
$\Delta^{*}_{t,i,d}$, its coordinate-wise crossover mask
$C^{*}_{t,i,d}$, and a bound-repair mask. Randomly repaired coordinates are
excluded because their values no longer represent the teacher mutation.

Let
\begin{equation}
  \widehat{\Delta}_{t,i,d}
  =
  v_{t,i,d}
  -
  x_{t,i,d}
\end{equation}
be the predicted dense mutation displacement. The operator imitation loss is
a coordinate-averaged Smooth L1 loss:
\begin{equation}
  \mathcal{L}_{\mathrm{op}}
  =
  \frac{1}{|\Omega|}
  \sum_{(i,d)\in\Omega}
  \operatorname{SmoothL1}_{\beta=0.02}
  \left(
    \widehat{\Delta}_{t,i,d}
    -
    \Delta^{*}_{t,i,d}
  \right),
\end{equation}
where $\Omega$ contains valid, teacher-crossover coordinates that were not
randomly repaired.

An auxiliary loss supervises the post-crossover proposal:
\begin{equation}
  \mathcal{L}_{\mathrm{cand}}
  =
  \frac{1}{|\Omega_{\mathrm{valid}}|}
  \sum_{(i,d)\in\Omega_{\mathrm{valid}}}
  \operatorname{SmoothL1}_{\beta=0.02}
  \left[
    (u_{t,i,d}-x_{t,i,d})
    -
    C^{*}_{t,i,d}\Delta^{*}_{t,i,d}
  \right].
\end{equation}
The complete supervised objective is
\begin{equation}
  \mathcal{L}_{\mathrm{sup}}
  =
  \mathcal{L}_{\mathrm{op}}
  +
  0.25\mathcal{L}_{\mathrm{cand}}.
  \label{eq:supervised_objective}
\end{equation}

\subsubsection{Online Meta-Training}
\label{app:online_meta_training}

The online stage jointly trains on continuous POM TF1--TF5 at
$D\in\{20,50,100,200\}$. Every Adam update accumulates equally weighted
gradients from all four dimensions.

For one population transition, the loss is
\begin{equation}
  \mathcal{L}_{\mathrm{online}}
  =
  \frac{
    \operatorname{mean}(\mathbf{y}_{t+1})
    -
    \operatorname{mean}(\mathbf{y}_{t})
  }{
    \max(
      |\operatorname{mean}(\mathbf{y}_{t})|,
      10^{-8}
    )
  }
  -
  \lambda_{\mathrm{div}}
  \frac{1}{D}
  \sum_{d=1}^{D}
  \operatorname{std}_{i}
  \left(
    x_{t+1,i,d}
  \right),
  \label{eq:online_metagbt_objective}
\end{equation}
where $\lambda_{\mathrm{div}}=0.005$.

The online-only model starts from random weights. Each full two-stage model
loads the supervised model weights but constructs a fresh Adam optimizer; the
supervised optimizer state is never reused. 

\subsection{Architecture and Hyperparameters}
\label{app:architecture_hyperparameters}

Table~\ref{tab:history_pom_architecture} summarizes the shared architecture.
The model has no fixed maximum decision dimension because all coordinate
parameters are shared before symmetric aggregation.

\begin{table*}[htbp]
  \centering
  \caption{Architecture of the continuous and mixed-variable A2M model.}
  \label{tab:history_pom_architecture}
  \small
  \setlength{\tabcolsep}{7pt}
  \begin{tabular}{lll}
    \toprule
    Component & Hyperparameter & Value \\
    \midrule
    Input & Population size $N$ & 100 \\
    Input & History length $K$ & 5 \\
    Input & Fitness features per individual & 2 \\
    Input & Coordinate features per coordinate & 3 \\
    Input & Fixed maximum decision dimension & None \\
    Coordinate encoder & Shared MLP & $3\rightarrow32\rightarrow32$ \\
    Coordinate encoder & Activation & GELU \\
    Coordinate encoder & Feature clipping & $[-8,8]$ \\
    Coordinate encoder & Symmetric pooling & Mean and RMS \\
    Coordinate projection & Projection & $64\rightarrow256$ \\
    Fusion & Population-position embedding & $100\times256$ \\
    Fusion & Dynamic gate initialization & 0.5 \\
    Fusion & Gate granularity & Per token and channel \\
    Transformer & Hidden width & 256 \\
    Transformer & Encoder blocks & 4 \\
    Transformer & Future decoder blocks & 3 \\
    Transformer & Attention heads & 4 \\
    Transformer & Attention head width & 64 \\
    Transformer & Feed-forward width & 1024 \\
    Transformer & Temporal encoding & RoPE, base $10{,}000$ \\
    Transformer & LayerScale initialization & 0.1 \\
    Transformer & Activation / dropout & GELU / 0 \\
    Transformer & Normalization & Pre-norm + final LayerNorm \\
    Operator & Operator heads / head width & $4/64$ \\
    Operator & Query/key activation & $\tanh$ \\
    Operator & Interaction coefficients & Signed $\tanh$ weights \\
    Operator & Proposal construction &
Signed population attention over the current population \\
    Operator & Structural regularization & Stochastic off-diagonal mask; self-interactions always active \\
    Crossover & Input / hidden width & $3/4$ \\
    Crossover & Decision rule & Hard Gumbel--Softmax \\
    Crossover & Temperature & 1.0 \\
    Selection & Rule & Fixed one-to-one greedy \\
    Parameters & Total trainable parameters & 7,953,089 \\
    \bottomrule
  \end{tabular}
\end{table*}

\begin{table*}[htbp]
  \centering
  \caption{Training configuration for continuous and mixed-variable models.}
  \label{tab:history_pom_training}
  \small
  \setlength{\tabcolsep}{6pt}
  \begin{tabular}{lll}
    \toprule
    Stage & Hyperparameter & Value \\
    \midrule
    Supervision & Continuous teachers &
    GTDE, GP-LogEI-GA, Projected-Lion \\
    Supervision & Mixed-variable teachers & GTDE \\
    Supervision & Training dimensions & 20, 50, 100, 200 \\
    Supervision & Teacher repeat & R0 \\
    Supervision & Continuous trajectories / windows & 6,480 / 615,600 \\
    Supervision & Mixed trajectories / windows & 2,160 / 205,200 \\
    Pretraining & GPUs / batch per GPU & $4/640$ \\
    Pretraining & Global batch size & 2,560 \\
    Pretraining & Epochs & 1 \\
    Pretraining & Optimizer & AdamW \\
    Pretraining & Learning rate & $8\times10^{-5}$ \\
    Pretraining & Weight decay & $10^{-4}$ \\
    Pretraining & Warm-up steps & 20 \\
    Pretraining & Minimum learning-rate ratio & 0.1 \\
    Pretraining & Gradient-norm clipping & 1.0 \\
    Pretraining & Precision & bfloat16 \\
    Online training & Functions & TF1--TF5 \\
    Online training & Joint dimensions & 20, 50, 100, 200 \\
    Online training & Dimension weights & $1:1:1:1$ \\
    Online training & GPUs & 8 \\
    Online training & Global tasks per function & 100 \\
    Online training & Optimizer & Fresh Adam \\
    Online training & Learning rate & $10^{-4}$ \\
    Online training & Diversity coefficient & 0.005 \\
    Online training & Gradient-norm clipping & 10.0 \\
    Online training & Precision & bfloat16 \\
    \bottomrule
  \end{tabular}
\end{table*}

\clearpage
\newpage

\section{MV-BBOB Test Suite}
We build mixed-variable optimization benchmarks by transforming the original 24 continuous BBOB functions. For each function index $\text{FID} \in \{1,2,\dots,24\}$, the construction steps are described in detail below.

First, we predefine a fixed integer-variable ratio $r_{\text{FID}}$ for each FID, as listed in Table~\ref{tab:fid_ratio}. The ratio controls the fraction of dimensions treated as integer variables. Given problem dimension $D$, the number of integer dimensions is calculated as
\[
k = \operatorname{round}(r_{\text{FID}} \cdot D).
\]
To guarantee mixed-variable characteristics, $k$ is clamped to satisfy $1 \le k \le D-1$, so that both integer and continuous variables exist simultaneously.

\begin{table}[htbp]
\centering
\caption{Predefined integer variable ratio $r_{\text{FID}}$ for each BBOB function}
\label{tab:fid_ratio}
\begin{tabular}{cccccc}
\hline
FID & $r_{\text{FID}}$ & FID & $r_{\text{FID}}$ & FID & $r_{\text{FID}}$ \\
\hline
1 & 0.70 & 9 & 0.15 & 17 & 0.40 \\
2 & 0.50 & 10 & 0.20 & 18 & 0.30 \\
3 & 0.40 & 11 & 0.40 & 19 & 0.20 \\
4 & 0.40 & 12 & 0.30 & 20 & 0.20 \\
5 & 0.70 & 13 & 0.60 & 21 & 0.15 \\
6 & 0.20 & 14 & 0.60 & 22 & 0.15 \\
7 & 0.60 & 15 & 0.20 & 23 & 0.40 \\
8 & 0.20 & 16 & 0.40 & 24 & 0.40 \\
\hline
\end{tabular}
\end{table}

Second, we select $k$ distinct dimension indices to be integer variables. To avoid integer dimensions clustering in partial positions, we adopt an equal-interval sampling strategy:
\begin{enumerate}
\item Generate initial candidate indices $\{0,\lfloor D/k\rfloor,2\lfloor D/k\rfloor,\dots\}$;
\item Remove duplicated indices and sort the set;
\item If the number of selected indices is less than $k$, sequentially supplement unused dimensions until exactly $k$ unique indices are obtained.
\end{enumerate}
The selected indices correspond to integer variables, and all remaining dimensions are continuous variables.

Before each FE, a candidate solution from the continuous search space is preprocessed for the original BBOB test instance. For each integer dimension, the value is rounded to integer grid points and clipped within variable bounds. Continuous dimensions remain unchanged. The modified vector is evaluated by the original BBOB test instance to obtain the objective value.

The resulting 24 mixed-variable test instances fully inherit continuous landscape characteristics, such as multimodality, separability, ill-conditioning and basin topology, as well as the original search bounds. The core difference lies in the feasible domain: a subset of decision variables is restricted to discrete integer grids via rounding projection before each FE, which shrinks the set of feasible solutions. Although the underlying function surface stays identical, the global optima of MV-BBOB instances generally deviates from the continuous optima of the original BBOB test instances. They coincide only when the continuous optimum locates exactly on the integer grid; in most scenarios, the mixed-variable optimum corresponds to the best feasible grid point nearby and yields an inferior objective value compared with the original continuous optimum.

\newpage
\clearpage

\section{Supplementary Results}
\begin{table*}[htbp]
  \centering
  \caption{Results on Continuous Optimization Scenario:  BBOB test suite.}
    \resizebox{15cm}{!}{
\begin{tabular}{c|c|ccccccccccc}
    \hline
    \multicolumn{1}{c|}{} & \textit{D} & GTDE  & sDMS-PSO & SAHLPSO & EvoTF & POM   & MetaDistill & LLM-Generate & EoH   & MEoH  & ReEvo & A2M \bigstrut\\
    \hline
    \multicolumn{1}{c|}{\multirow{5}[2]{*}{F1}} & 20    & 6.59E-02(2.84E-02) & 6.92E+01(1.40E+01) & 2.08E+01(6.34E+00) & 3.78E-01(8.52E-02) & 6.29E-12(4.14E-12) & 1.43E-01(6.77E-02) & 3.73E+01(7.37E+00) & 6.19E+00(1.08E+00) & 6.21E+00(8.91E-01) & 7.16E+01(1.46E+01) & \textbf{8.26E-23(9.11E-23)} \bigstrut[t]\\
    \multicolumn{1}{c|}{} & 50    & 4.64E-01(1.49E-01) & 2.68E+02(5.55E+01) & 1.16E+02(2.03E+01) & 1.53E+01(2.31E+00) & 2.24E-11(8.23E-12) & 4.02E-01(8.37E-02) & 4.29E+02(5.34E+01) & 2.27E+00(2.21E-01) & 5.33E+01(5.20E+00) & 5.32E+02(7.39E+01) & \textbf{4.57E-21(2.19E-21)} \\
    \multicolumn{1}{c|}{} & 100   & 2.40E+00(4.63E-01) & 6.59E+02(1.10E+02) & 2.66E+02(4.69E+01) & 9.51E+01(1.83E+01) & 3.89E-11(2.03E-11) & 8.74E-01(1.25E-01) & 1.58E+03(6.39E+01) & 1.02E+01(1.07E+00) & 2.45E+02(1.59E+01) & 1.60E+03(2.06E+02) & \textbf{8.13E-20(4.38E-20)} \\
    \multicolumn{1}{c|}{} & 200   & 2.36E+01(3.32E+00) & 1.35E+03(1.40E+02) & 6.67E+02(6.76E+01) & 6.52E+02(1.09E+02) & 8.93E-11(3.72E-11) & 1.96E+00(2.54E-01) & 4.20E+03(1.37E+02) & 5.01E+01(3.55E+00) & 1.18E+03(7.75E+01) & 4.01E+03(3.81E+02) & \textbf{1.21E-18(8.69E-19)} \\
    \multicolumn{1}{c|}{} & 500   & 5.46E+02(4.33E+01) & 3.89E+03(6.53E+02) & 1.94E+03(2.61E+02) & 4.97E+03(3.88E+02) & 1.57E-10(6.48E-11) & 5.37E+00(4.09E-01) & 1.32E+04(2.32E+02) & 4.67E+02(2.38E+01) & 7.01E+03(1.62E+02) & 9.55E+03(9.71E+02) & \textbf{3.28E-17(1.43E-17)} \bigstrut[b]\\
    \hline
    \multicolumn{1}{c|}{\multirow{5}[2]{*}{F2}} & 20    & 7.74E-04(4.49E-04) & 2.72E-01(1.28E-01) & 2.97E-01(1.58E-01) & 2.91E-03(1.51E-03) & 1.73E-13(1.20E-13) & 2.83E-03(2.76E-03) & 2.78E-01(7.01E-02) & 1.53E-01(6.30E-02) & 7.72E-02(3.53E-02) & 6.16E-01(2.53E-01) & \textbf{6.88E-21(6.85E-21)} \bigstrut[t]\\
\hline
    \end{tabular}}
  \label{tab:continuous_bbob}
\end{table*}

\begin{table*}[htbp]
  \centering
  \caption{Results on Continuous Optimization Scenario:  CEC2020 and CEC2026 test suites.}
    \resizebox{18cm}{!}{\begin{tabular}{cc|ccccccccccc}
    \hline
    \multicolumn{1}{c|}{} & \textit{D} & GTDE  & sDMS-PSO & SAHLPSO & EvoTF & POM   & MetaDistill & LLM-Generate & EoH   & MEoH  & ReEvo & A2M \bigstrut\\
    \hline
    \multicolumn{1}{c|}{\multirow{2}[2]{*}{CEC2020-F1}} & 10    & \textbf{4.08E+06(2.39E+06)} & 4.53E+09(1.91E+09) & 2.07E+08(2.25E+08) & 9.20E+06(6.13E+06) & 4.33E+07(8.40E+07) & 5.50E+09(1.45E+09) & 3.95E+08(9.42E+07) & 1.40E+09(3.64E+08) & 8.44E+07(2.24E+07) & 9.51E+08(3.62E+08) & 1.48E+07(4.34E+07) \bigstrut[t]\\
    \multicolumn{1}{c|}{} & 20    & \textbf{1.01E+07(6.00E+06)} & 2.62E+10(4.46E+09) & 2.00E+09(1.80E+09) & 2.62E+08(1.94E+08) & 1.45E+09(1.04E+09) & 2.12E+10(2.40E+09) & 4.82E+09(1.31E+09) & 6.80E+08(1.23E+08) & 6.56E+08(1.08E+08) & 6.64E+09(1.93E+09) & 5.47E+08(5.67E+08) \bigstrut[b]\\
    \hline
    \multicolumn{1}{c|}{\multirow{2}[2]{*}{CEC2020-F2}} & 10    & \textbf{1.76E+03(2.92E+02)} & 2.41E+03(2.71E+02) & 2.45E+03(5.57E+02) & 2.87E+03(3.83E+02) & 2.13E+03(2.44E+02) & 2.83E+03(1.78E+02) & 2.64E+03(1.43E+02) & 2.80E+03(1.77E+02) & 2.37E+03(2.22E+02) & 2.69E+03(1.95E+02) & 1.85E+03(2.86E+02) \bigstrut[t]\\
    \multicolumn{1}{c|}{} & 20    & \textbf{1.96E+03(3.23E+02)} & 4.98E+03(4.42E+02) & 5.09E+03(1.14E+03) & 4.17E+03(5.33E+02) & 4.01E+03(6.71E+02) & 6.07E+03(3.60E+02) & 5.54E+03(1.98E+02) & 5.49E+03(4.90E+02) & 5.09E+03(4.50E+02) & 5.51E+03(3.61E+02) & 3.63E+03(8.74E+02) \bigstrut[b]\\
    \hline
    \multicolumn{1}{c|}{\multirow{2}[2]{*}{CEC2020-F3}} & 10    & 7.40E+02(9.47E+00) & 8.12E+02(2.48E+01) & 7.70E+02(2.00E+01) & 1.38E+03(2.06E+02) & 7.31E+02(4.84E+00) & 8.17E+02(1.39E+01) & 7.87E+02(1.03E+01) & 8.14E+02(1.51E+01) & 7.61E+02(8.37E+00) & 8.29E+02(1.49E+01) & \textbf{7.29E+02(7.79E+00)} \bigstrut[t]\\
\hline
    \end{tabular}}
  \label{tab:continuous_cec}
\end{table*}

\begin{table*}[htbp]
  \centering
  \caption{Results on Mixed-Variable Optimization Scenario: MV-BBOB Test Suite.}
    \resizebox{16cm}{!}{\begin{tabular}{c|c|ccccccccccc}
    \hline
          & \textit{D} & GTDE  & DE$_{\textup{MV}}$ & PSO$_{\textup{MV}}$ & EvoTF & POM   & MetaDistill & LLM-Generate & EoH   & MEoH  & ReEvo & A2M \bigstrut\\
    \hline
    \multirow{5}[2]{*}{F1} & 20    & 2.51E-02(3.23E-02) & 1.81E+02(2.39E+01) & 5.80E+01(1.97E+01) & 2.91E-01(1.87E-01) & 3.30E-14(2.96E-14) & 2.85E-02(2.24E-02) & 1.54E+02(2.73E+01) & 1.94E-03(5.03E-04) & 1.55E-04(1.51E-04) & 2.85E-04(2.34E-04) & \textbf{1.01E-21(2.07E-21)} \bigstrut[t]\\
          & 50    & 1.17E-01(8.81E-02) & 9.12E+02(1.43E+02) & 3.56E+02(7.75E+01) & 1.82E+01(2.45E+00) & 3.22E-12(2.32E-12) & 1.01E-01(3.52E-02) & 9.93E+02(6.91E+01) & 4.72E-03(1.16E-03) & 4.26E+00(1.76E+00) & 6.49E-03(2.98E-03) & \textbf{2.58E-18(2.37E-18)} \\
          & 100   & 9.62E-01(6.62E-01) & 2.45E+03(2.59E+02) & 1.30E+03(2.38E+02) & 1.09E+02(2.08E+01) & 1.13E-11(4.60E-12) & 2.12E-01(6.41E-02) & 2.39E+03(1.91E+02) & 1.08E-02(1.78E-03) & 6.66E+01(9.37E+00) & 2.38E-01(5.20E-01) & \textbf{2.48E-17(1.49E-17)} \\
          & 200   & 2.27E+01(4.26E+00) & 5.62E+03(1.53E+02) & 3.40E+03(4.82E+02) & 7.16E+02(1.15E+02) & 3.15E-11(1.08E-11) & 5.32E-01(7.76E-02) & 5.74E+03(4.86E+02) & 2.05E-02(1.94E-03) & 4.49E+02(3.82E+01) & 1.58E+01(4.04E+00) & \textbf{1.49E-16(1.58E-16)} \\
          & 500   & 5.52E+02(3.33E+01) & 1.49E+04(2.86E+02) & 1.14E+04(1.04E+03) & 5.20E+03(4.01E+02) & 5.40E-11(1.51E-11) & 1.53E+00(2.42E-01) & 1.61E+04(3.06E+02) & 3.03E+00(3.50E+00) & 4.11E+03(1.74E+02) & 3.10E+02(2.45E+01) & \textbf{2.03E-16(1.45E-16)} \bigstrut[b]\\
    \hline
    \multirow{5}[2]{*}{F2} & 20    & 8.05E-04(1.09E-03) & 9.80E-02(2.70E-02) & 1.02E+00(7.70E-01) & 2.97E-03(1.53E-03) & 1.62E-14(1.41E-14) & 1.80E-03(2.53E-03) & 9.56E-02(2.15E-02) & 2.27E-04(1.70E-04) & 1.40E-03(1.89E-03) & 1.77E-03(1.23E-03) & \textbf{8.26E-21(2.16E-20)} \bigstrut[t]\\
\hline
    \end{tabular}}
\end{table*}
\newpage
\clearpage
\begin{table*}[htbp]
  \centering
  \caption{Results on Mixed-Variable Optimization Scenario: EOPCCV Test Suite.}
    \resizebox{18cm}{!}{\begin{tabular}{c|c|ccccccccccc}
    \hline
          & \textit{D} & GTDE  & DE$_{\textup{MV}}$ & PSO$_{\textup{MV}}$ & EvoTF & POM   & MetaDistill & LLM-Generate & EoH   & MEoH  & ReEvo & A2M \bigstrut\\
    \hline
    F1    & 10    & 1.89E+01(3.02E+01) & 1.22E+02(3.88E+01) & 1.21E+03(6.19E+02) & 1.44E+02(1.50E+02) & 8.20E+01(3.98E+01) & 4.29E+03(8.54E+02) & 9.32E+01(4.20E+01) & \textbf{1.60E+01(2.50E+01)} & 1.09E+02(1.07E+02) & 1.20E+02(1.33E+02) & 1.84E+01(2.12E+01) \bigstrut[t]\\
    F2    & 10    & 5.46E+01(1.15E+01) & 4.77E+01(5.89E+00) & 6.47E+01(1.11E+01) & 1.60E+02(4.64E+01) & \textbf{1.87E+01(7.69E+00)} & 9.11E+01(1.27E+01) & 4.79E+01(7.36E+00) & 3.67E+01(9.85E+00) & 3.40E+01(1.28E+01) & 6.59E+01(1.80E+01) & 1.95E+01(9.40E+00) \\
    F3    & 10    & 3.46E+00(1.36E+00) & 6.38E+00(8.10E-01) & 1.24E+01(2.58E+00) & 1.90E+01(3.00E+00) & 4.40E+00(1.08E+00) & 1.56E+01(6.84E-01) & 5.80E+00(6.38E-01) & 1.85E+00(1.39E+00) & 3.83E+00(2.05E+00) & 4.24E+00(1.99E+00) & \textbf{1.71E+00(1.06E+00)} \\
    F4    & 10    & 5.23E+00(8.54E+00) & 5.25E-01(2.06E-01) & 2.11E+01(1.13E+01) & 1.30E+01(2.03E+01) & 4.13E-01(7.88E-01) & 7.66E+01(1.30E+01) & 3.76E-01(1.40E-01) & \textbf{2.55E-03(1.30E-03)} & 1.69E+00(4.12E+00) & 2.48E+01(2.02E+01) & 6.64E-02(1.35E-01) \\
    F5    & 10    & 3.18E+00(3.22E+00) & 1.89E+00(4.48E-01) & 1.28E+01(1.05E+01) & 2.86E+00(2.81E+00) & 1.41E+00(5.48E-01) & 3.84E+01(5.33E+00) & 1.75E+00(3.46E-01) & 9.66E-01(3.33E-01) & \textbf{3.85E-01(1.41E-01)} & 6.20E+00(3.03E+00) & 7.95E-01(1.08E-01) \\
    F6    & 10    & 2.34E+01(3.42E+01) & 1.42E+02(4.76E+01) & 1.04E+03(6.41E+02) & 3.86E+02(1.00E+03) & 8.12E+01(3.35E+01) & 4.27E+03(4.93E+02) & 1.24E+02(3.96E+01) & 4.84E+01(3.54E+01) & \textbf{2.91E-01(1.26E+00)} & 9.22E+02(1.28E+03) & 1.90E+01(1.79E+01) \\
\hline
    \end{tabular}}
\end{table*}
\newpage
\clearpage
\newpage
\clearpage
\begin{table*}[htbp]
  \centering
  \caption{Results on Expensive Optimization Scenario: BBOB Test Suite.}
    \resizebox{18cm}{!}{\begin{tabular}{p{5.18em}|c|ccccccccccc}
    \hline
    \multicolumn{1}{c|}{} & \textit{D} & GTDE  & SADE-AMSS & GP-LogEI & EvoTF & POM   & MetaDistill & LLM-Generate & EoH   & MEoH  & ReEvo & A2M \bigstrut\\
    \hline
    \multicolumn{1}{c|}{\multirow{5}[2]{*}{F1}} & 20    & 1.89E+02(4.67E+01) & 5.04E+00(2.65E+00) & 1.73E+02(2.11E+01) & 4.39E+02(1.02E+02) & 1.84E+01(7.24E+00) & 3.30E+00(1.02E+00) & 3.06E+02(2.51E+01) & 3.24E+02(2.59E+01) & 3.56E+02(3.88E+01) & 2.29E+02(4.58E+01) & \textbf{8.92E-01(3.56E-01)} \bigstrut[t]\\
    \multicolumn{1}{c|}{} & 50    & 6.57E+02(1.51E+02) & 1.67E+02(6.91E+01) & 8.26E+02(5.41E+01) & 1.34E+03(1.90E+02) & 1.36E+02(2.10E+01) & 3.21E+01(1.58E+01) & 1.09E+03(7.41E+01) & 1.08E+03(9.53E+01) & 1.17E+03(9.31E+01) & 7.65E+02(1.45E+02) & \textbf{5.71E+00(1.49E+00)} \\
    \multicolumn{1}{c|}{} & 100   & 1.47E+03(1.32E+02) & 6.97E+02(1.81E+02) & 2.14E+03(7.95E+01) & 2.83E+03(2.73E+02) & 3.75E+02(5.38E+01) & 1.63E+02(3.56E+01) & 2.58E+03(8.39E+01) & 2.50E+03(2.12E+02) & 2.29E+03(4.93E+02) & 2.11E+03(3.10E+02) & \textbf{2.05E+01(4.10E+00)} \\
    \multicolumn{1}{c|}{} & 200   & 3.69E+03(3.35E+02) & 2.96E+03(2.90E+02) & 5.47E+03(1.64E+02) & 5.84E+03(4.61E+02) & 9.42E+02(8.57E+01) & 4.97E+02(7.33E+01) & 5.57E+03(1.49E+02) & 5.29E+03(4.77E+02) & 5.14E+03(9.55E+02) & 4.56E+03(1.07E+03) & \textbf{5.09E+01(1.14E+01)} \\
    \multicolumn{1}{c|}{} & 500   & 1.06E+04(3.33E+02) & 7.79E+03(3.86E+02) & 1.48E+04(2.30E+02) & 1.54E+04(7.55E+02) & 2.76E+03(1.19E+02) & 1.65E+03(1.29E+02) & 1.49E+04(2.79E+02) & 1.33E+04(1.28E+03) & 1.20E+04(3.48E+03) & 1.45E+04(1.11E+03) & \textbf{1.63E+02(2.08E+01)} \bigstrut[b]\\
    \hline
    \multicolumn{1}{c|}{\multirow{5}[2]{*}{F2}} & 20    & 2.46E+00(1.58E+00) & 8.01E-01(2.82E-01) & 7.40E-01(1.79E-01) & 1.14E+01(7.15E+00) & 2.19E-01(1.48E-01) & 4.07E-02(2.21E-02) & 2.63E+00(8.04E-01) & 2.87E+00(1.23E+00) & 4.27E+00(1.26E+00) & 1.58E+00(7.81E-01) & \textbf{9.60E-03(7.92E-03)} \bigstrut[t]\\
\hline
    \end{tabular}}
\end{table*}
\newpage
\clearpage
\begin{table*}[htbp]
  \centering
  \caption{Results on Expensive Optimization Scenario: CEC2020 and CEC2026 Test Suites.}
    \resizebox{18cm}{!}{\begin{tabular}{c|c|ccccccccccc}
    \hline
          & \textit{D} & GTDE  & SADE-AMSS & GP-LogEI & EvoTF & POM   & MetaDistill & LLM-Generate & EoH   & MEoH  & ReEvo & A2M \bigstrut\\
    \hline
    \multirow{2}[2]{*}{CEC2020-F1} & 10    & 7.88E+09(3.52E+09) & 2.01E+09(1.32E+09) & 2.64E+09(6.64E+08) & 6.35E+10(2.85E+10) & 4.58E+09(1.89E+09) & 1.24E+10(2.92E+09) & 1.04E+10(3.98E+09) & 1.06E+10(3.52E+09) & 1.26E+10(2.88E+09) & 2.17E+09(1.54E+09) & \textbf{8.63E+08(5.27E+08)} \bigstrut[t]\\
          & 20    & 2.86E+10(8.87E+09) & 1.02E+10(3.30E+09) & 2.02E+10(2.02E+09) & 1.26E+11(4.53E+10) & 1.69E+10(3.48E+09) & 3.71E+10(4.08E+09) & 4.52E+10(9.17E+09) & 4.82E+10(1.04E+10) & 4.80E+10(7.22E+09) & 1.19E+10(7.38E+09) & \textbf{7.16E+09(2.41E+09)} \bigstrut[b]\\
    \hline
    \multirow{2}[2]{*}{CEC2020-F2} & 10    & 3.16E+03(3.98E+02) & \textbf{2.98E+03(2.18E+02)} & 3.26E+03(2.12E+02) & 4.20E+03(5.89E+02) & 3.25E+03(2.03E+02) & 3.32E+03(3.25E+02) & 3.31E+03(1.81E+02) & 3.34E+03(2.04E+02) & 3.44E+03(2.15E+02) & 3.24E+03(2.38E+02) & 3.13E+03(2.11E+02) \bigstrut[t]\\
          & 20    & 5.80E+03(5.68E+02) & \textbf{5.49E+03(3.74E+02)} & 6.43E+03(3.31E+02) & 7.44E+03(7.30E+02) & 6.42E+03(3.83E+02) & 6.64E+03(2.91E+02) & 6.44E+03(2.77E+02) & 6.56E+03(3.72E+02) & 6.49E+03(3.37E+02) & 6.33E+03(4.46E+02) & 6.14E+03(2.91E+02) \bigstrut[b]\\
    \hline
    \multirow{2}[2]{*}{CEC2020-F3} & 10    & 9.51E+02(6.78E+01) & 8.09E+02(2.60E+01) & 8.68E+02(1.91E+01) & 1.75E+03(3.79E+02) & 8.05E+02(1.78E+01) & 8.59E+02(1.64E+01) & 1.04E+03(6.48E+01) & 1.06E+03(7.26E+01) & 1.14E+03(5.07E+01) & 1.01E+03(6.19E+01) & \textbf{7.74E+02(8.31E+00)} \bigstrut[t]\\
          & 20    & 1.57E+03(1.86E+02) & 1.08E+03(1.32E+02) & 1.43E+03(4.78E+01) & 3.23E+03(5.21E+02) & 1.02E+03(3.41E+01) & 1.12E+03(3.39E+01) & 1.92E+03(1.53E+02) & 1.99E+03(1.43E+02) & 2.11E+03(1.86E+02) & 1.68E+03(1.38E+02) & \textbf{9.28E+02(1.78E+01)} \bigstrut[b]\\
\hline
    \end{tabular}}
\end{table*}
\newpage
\clearpage

\begin{table*}[htbp]
  \centering
  \caption{Ablation Study Results.}
    \resizebox{13cm}{!}{\begin{tabular}{cc|cccc|r|ccc}
    \hline
         & {\textit{D}} & {EC-based only (GTDE)} & {BO-based only (GP-LogEI)} & {Gradient-based only (Lion)} & {A2M (Full)} &       & {Imitation learning only} & {Online learning only} & {A2M (Full)} \bigstrut\\
    \hline
    \multicolumn{1}{c|}{\multirow{5}[2]{*}{F1}} & 20    & 3.34E+02(4.21E+01) & 2.32E+02(4.05E+01) & 2.07E-16(2.30E-16) & \textbf{8.26E-23(9.11E-23)} &       & 6.47E-05(6.24E-05) & 3.41E+02(4.08E+01) & \textbf{8.26E-23(9.11E-23)} \bigstrut[t]\\
    \multicolumn{1}{c|}{} & 50    & 1.16E+03(8.28E+01) & 1.16E+03(8.28E+01) & 1.76E-15(1.83E-15) & \textbf{4.57E-21(2.19E-21)} &       & 5.92E-04(6.34E-04) & 1.16E+03(8.28E+01) & \textbf{4.57E-21(2.19E-21)} \\
    \multicolumn{1}{c|}{} & 100   & 2.66E+03(1.13E+02) & 2.66E+03(1.13E+02) & 3.45E-15(3.27E-15) & \textbf{8.13E-20(4.38E-20)} &       & 1.62E-03(8.18E-04) & 2.66E+03(1.13E+02) & \textbf{8.13E-20(4.38E-20)} \\
    \multicolumn{1}{c|}{} & 200   & 5.69E+03(1.79E+02) & 5.69E+03(1.79E+02) & 6.90E-15(5.17E-15) & \textbf{1.21E-18(8.69E-19)} &       & 4.66E-03(2.10E-03) & 5.69E+03(1.79E+02) & \textbf{1.21E-18(8.69E-19)} \\
    \multicolumn{1}{c|}{} & 500   & 1.50E+04(2.32E+02) & 1.50E+04(2.32E+02) & 1.42E-14(1.30E-14) & \textbf{3.28E-17(1.43E-17)} &       & 2.05E-02(6.48E-03) & 1.50E+04(2.32E+02) & \textbf{3.28E-17(1.43E-17)} \bigstrut[b]\\
    \hline
    \multicolumn{1}{c|}{\multirow{5}[2]{*}{F2}} & 20    & 2.43E+00(7.19E-01) & 3.99E-01(1.18E-01) & 1.47E-17(2.96E-17) & \textbf{6.88E-21(6.85E-21)} &       & 5.27E-06(1.12E-05) & 1.95E+00(5.45E-01) & \textbf{6.88E-21(6.85E-21)} \bigstrut[t]\\
    \multicolumn{1}{c|}{} & 50    & 2.96E+01(5.18E+00) & 1.37E+01(1.94E+00) & 1.68E-16(1.04E-16) & \textbf{4.66E-18(6.08E-18)} &       & 4.64E-05(3.20E-05) & 2.85E+01(5.03E+00) & \textbf{4.66E-18(6.08E-18)} \\
\hline
    \end{tabular}}
\end{table*}

\clearpage

\bibliographystyle{IEEEtran}
\bibliography{myref}